\documentclass[11pt]{article}
\usepackage[final]{acl}

\usepackage{times}
\usepackage{latexsym}
\usepackage[T1]{fontenc}
\usepackage[utf8]{inputenc}
\usepackage{microtype}
\usepackage{graphicx}
\usepackage{amsmath}
\usepackage{enumitem}

\usepackage{amssymb}
\usepackage{mathtools}
\usepackage{amsthm}
\usepackage{multirow}
\usepackage{url}
\usepackage{algorithm}
\usepackage{algorithmic}
\usepackage{makecell}
\usepackage{tcolorbox}
\usepackage{booktabs}
\usepackage[table]{xcolor}
\title{Bridging What the Model Thinks and How It Speaks:\\
Expressive Speech Generation via Self-Aware Intent-Realization Alignment}

\author{
    \textbf{Kuang Wang}$^{1,2}$,\ 
    \textbf{Lai Wei}$^{1,2}$,\ 
    \textbf{Ping Lin}$^{1,3}$,\ 
    \textbf{Qibing Bai}$^{1,2}$,\ 
    \textbf{Wenkai Fang}$^{4}$,\ 
    \textbf{Li Zhou}$^{1}$, \\
    \textbf{Feng Jiang}$^{5}$,\ 
    \textbf{Zhongjie Jiang}$^{2}$,\ 
    \textbf{Jun Huang}$^{2}$,\ 
    \textbf{Yannan Wang}$^{2}$,\ 
    \textbf{Haizhou Li}$^{1}$ \\[4pt]
    $^{1}$The Chinese University of Hong Kong, Shenzhen \quad
    $^{2}$Tencent Ethereal Audio Lab \quad
    $^{3}$Southeast University \\
    $^{4}$Zhejiang University \quad
    $^{5}$Shenzhen University of Advanced Technology \\[4pt]
    \texttt{kuangwang@link.cuhk.edu.cn},\ \texttt{jiangfeng@suat-sz.edu.cn}
}

\begin{document}
\maketitle

\begin{abstract}
Speech Language Models (SLMs) exhibit strong semantic understanding, yet often fail to translate this capacity into expressive acoustic realization, producing speech with flattened prosody and misaligned emotion. We identify this mismatch as the \textbf{semantic understanding--acoustic realization gap}. Existing approaches typically rely on externally specified proxies, such as emotion labels or style prompts, which require annotations and struggle to capture dynamically evolving expressive intent throughout dialogue.
To overcome these limitations, we propose \textbf{SASLM}
(\textbf{S}elf-\textbf{A}ware \textbf{S}peech \textbf{L}anguage \textbf{M}odel), a proxy-free framework that bridges what the model thinks and how it speaks through self-aware intent--realization alignment:
(1) \textit{Intent-Aware Bridging} self-distills expressive intent from the model's own evolving semantic generation states via a Variational Information Bottleneck (VIB), thereby guiding expressive speech realization without external expressive supervision; while
(2) \textit{Realization-Aware Alignment} reflectively aligns generated acoustics with intended expression through self-reward optimization, progressively improving intent--realization consistency during speech generation.
Despite using only 3B parameters and 800 hours of expressive speech data, SASLM achieves state-of-the-art performance on EchoMind among open-source systems, surpassing models over 10$\times$ larger and approaching commercial systems.\footnote{Code is available at \url{https://github.com/wangkevin02/SASLM}, and demo samples can be found at \url{https://wangkevin02.github.io/SASLM/}.}

\end{abstract}

\section{Introduction}
\label{sec:intro}

\begin{figure}[t]
\centering
\includegraphics[width=0.48\textwidth]{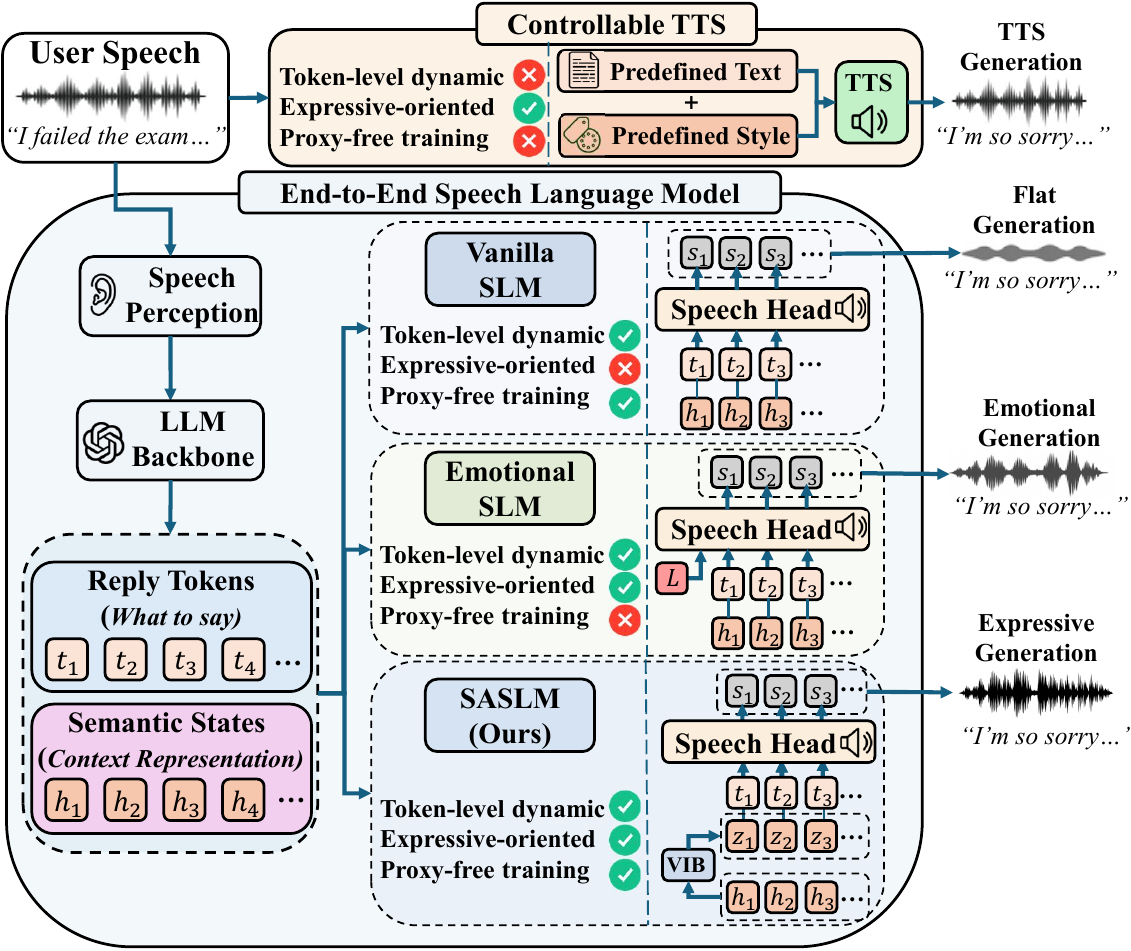}
\caption{
Comparison of context-aware expressive speech generation paradigms, categorized by their \textbf{expressive control signals}. Controllable TTS depends on external style references, whereas Vanilla SLMs naively reuse semantic hidden states $h_i$ to modulate text tokens $t_i$ into speech $s_i$. Unlike Emotional SLMs, which inject explicit emotion labels $L$ prior to synthesis, SASLM self-distills intrinsic expressive intent $z_i$ directly from $h_i$, eliminating proxy supervision.
}
\label{fig:intro_demo}
\end{figure}

Speech Language Models (SLMs) have substantially advanced multimodal interaction by integrating speech and text within unified generative frameworks, achieving remarkable success in multimodal understanding~\cite{xu2025qwen25omnitechnicalreport, xu2025qwen3omnitechnicalreport}. However, stronger semantic intelligence does not necessarily yield better spoken interaction. Unlike text generation, spoken communication depends not only on \textit{what} is said, but also on \textit{how} it is acoustically realized~\cite{zhu2022effects}, requiring contextually appropriate emotional intensity~\cite{scherer2003vocal}, prosody~\cite{ladd2008intonational}, and spontaneous naturalness~\cite{shriberg2005spontaneous}.

While large-scale training endows current vanilla SLMs~\cite{zeng2024glm, xu2025qwen25omnitechnicalreport} with robust semantic capabilities, they still struggle to translate this intelligence into context-aware expressive speech interaction. As illustrated in Fig.~\ref{fig:intro_demo}, they directly reuse text-prediction-oriented semantic hidden states for speech realization, which are not inherently aligned with expressive acoustic patterns and often yield semantically correct but prosody-flattened ``spoken text''~\cite{tu2025ultravoice, zhan2025vstyle, fang2025llama}. We refer to this mismatch as the \textbf{semantic understanding--acoustic realization gap}: semantic states that support strong understanding are not necessarily acoustically usable for expressive speech generation.


Existing approaches attempt to bridge this gap by externalizing expressive intent as explicit proxies. Cascaded systems~\cite{li2024style} typically use controllable TTS with style prompts inferred by speech understanding models, while emotional SLMs~\cite{chen2025emova, wang2025empathy, gao2025lucy} internalize it as affective proxies via joint emotion–speech prediction. Although effective in constrained scenarios, these approaches struggle to capture dynamically evolving intent and require costly annotations. Moreover, their training paradigms remain largely open-loop, optimizing token-level imitation without explicitly verifying cross-modal consistency between the intended expression and its acoustic realization.


To address these limitations, we rethink context-aware expressive speech generation as a problem of self-aware intent--realization alignment. 
To this end, we propose \textbf{SASLM}
(\textbf{S}elf-\textbf{A}ware \textbf{S}peech \textbf{L}anguage \textbf{M}odel), a proxy-free framework that intrinsically derives expressive intent from semantic generation dynamics, rather than relying on externally specified controls, and aligns generated acoustics with the intended expression via reflective self-feedback.
This reformulates the task as a closed-loop process that jointly models semantic understanding and acoustic realization. Concretely, SASLM realizes this self-aware loop via two coupled mechanisms.
\textit{Intent-Aware Bridging} internally forms expressive intent without external supervision by using a Variational Information Bottleneck (VIB) to self-distill evolving semantic states, disentangling acoustically usable intent governing \textit{how to speak} from lexical content already specified by generated tokens.
\textit{Realization-Aware Alignment} further enforces intent--realization consistency via a self-reward mechanism: SASLM acts as its own critic, scoring generated speech against its intended expression and optimizing itself using utterance-level rubric rewards.

Experiments on the EchoMind benchmark~\cite{zhou2025echomind} demonstrate that SASLM achieves state-of-the-art (SOTA) expressive performance among open-source models using only 3B parameters and 800 hours of expressive speech data, outperforming the $10\times$ larger Qwen3-Omni-30B while approaching closed-source commercial systems. Our main contributions are as follows:

    (1) We identify the \textbf{semantic understanding--acoustic realization gap} in current SLMs and reformulate expressive speech generation as a problem of self-aware intent--realization alignment.
    
    (2) We propose SASLM, a proxy-free framework that bridges semantic understanding and expressive acoustic realization through \textit{Intent-Aware Bridging} and \textit{Realization-Aware Alignment}.
    
    (3) Extensive experiments show that SASLM achieves SOTA expressive-speech performance among open-source SLMs with only 3B parameters and 800 hours of expressive speech data.

\section{Related Work}

\begin{figure}[t]
\centering
\includegraphics[width=\columnwidth]{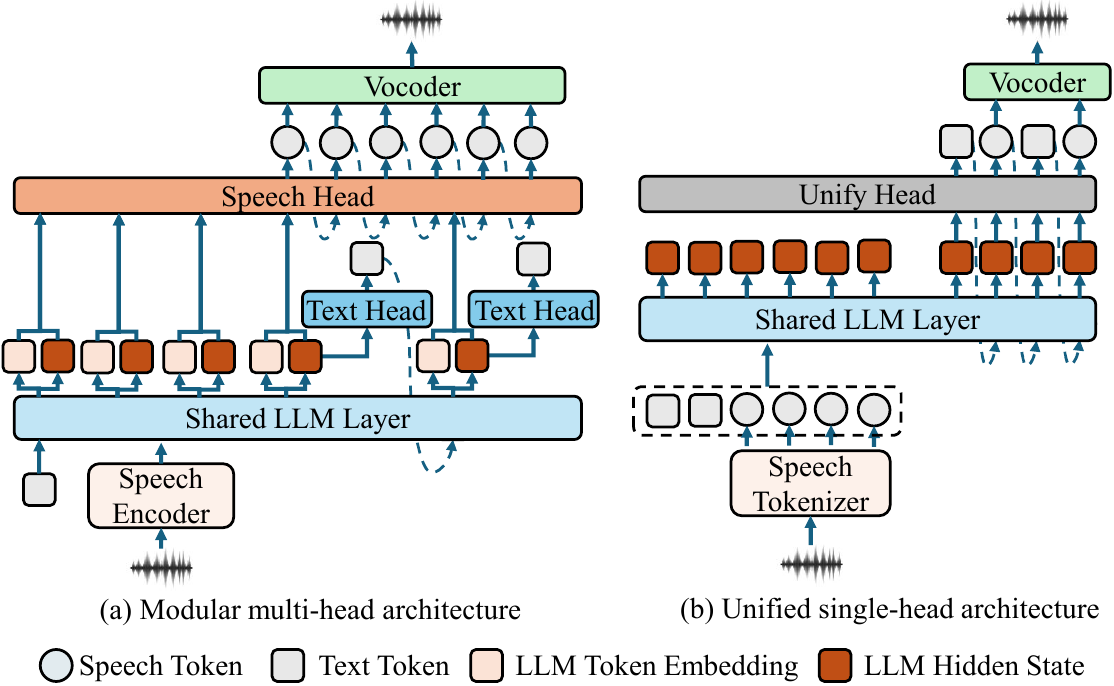}
\caption{Illustration of existing end-to-end SLM architectures: (a) Modular multi-head architecture; (b) Unified single-head architecture.}
\label{fig:diff_arch_slm}
\end{figure}

\subsection{Semantic-to-Speech Generation in SLMs}
\label{sec:related_work_slm}

Existing SLMs generally adopt two semantic-to-speech generation paradigms (Fig.~\ref{fig:diff_arch_slm}). Modular multi-head models\cite{xu2025qwen25omnitechnicalreport, fang2025llama, team2025fun} use a shared LLM backbone with dedicated speech encoders and generation heads, directly reusing LLM hidden states for acoustic generation. Unified single-head models\cite{zeng2024glm, nguyen2025spirit, maimon2025scaling} extend the LLM vocabulary with speech tokens and generate interleaved text–speech sequences autoregressively. Despite architectural differences, both paradigms mainly repurpose text-prediction-oriented semantic states for speech generation, without explicitly modeling acoustically usable expressive intent.

\subsection{Proxy-based Expressive Speech Generation}
\label{sec:related_work_expressive}

Early work on expressive speech generation mainly focused on text-to-speech (TTS) systems~\cite{du2024cosyvoice, anastassiou2024seed}, where large-scale pretraining enables rich acoustic continuation conditioned on human-written text and predefined style prompts. These systems typically rely on externally specified controls, such as speaking styles or emotion descriptions, to guide acoustic realization.

More recently, expressive generation has been extended to SLMs. Lucy~\cite{gao2025lucy} and EMOVA~\cite{chen2025emova} employ discrete emotion tokens to control speech generation, while Empathy Omni~\cite{wang2025empathy} introduces auxiliary emotion objectives to align generated speech with users' emotional states. Nevertheless, these methods still rely on externally proxy supervision, rather than dynamically modeling expressive intent as the dialogue unfolds in  proxy-free manner.

\subsection{Alignment for Speech Generation}
\label{sec:related_work_rl}

Recent reinforcement learning advances~\cite{guo2025deepseek}, such as RLHF~\cite{Ouyang0JAWMZASR22}, AI feedback~\cite{lee2023rlaif}, self-rewarding~\cite{fang2025serl}, and rubric-based reward modeling~\cite{gunjal2025rubrics}, have not only enhanced text generation but also extended alignment techniques to the speech domain.

Existing text-to-speech systems commonly use preference optimization methods such as DPO~\cite{rafailov2023direct, du2024cosyvoice, gao2025emo}, which rely on costly human-annotated preference pairs. Recent work further explores online reinforcement learning for speech generation, including reward modeling for acoustic quality~\cite{gao2025differentiable, gao2025explore}. In SLMs, SI-SDA~\cite{wang2025self} leverages saliency-based pseudo-labels to improve speech understanding. However, these methods mainly target utterance-level TTS quality or speech-understanding ability, leaving semantic intent--expressive realization alignment in SLMs largely underexplored.

\section{Self-Aware Speech Language Model}

\begin{figure*}[t]
\centering
\includegraphics[width=0.85\textwidth]{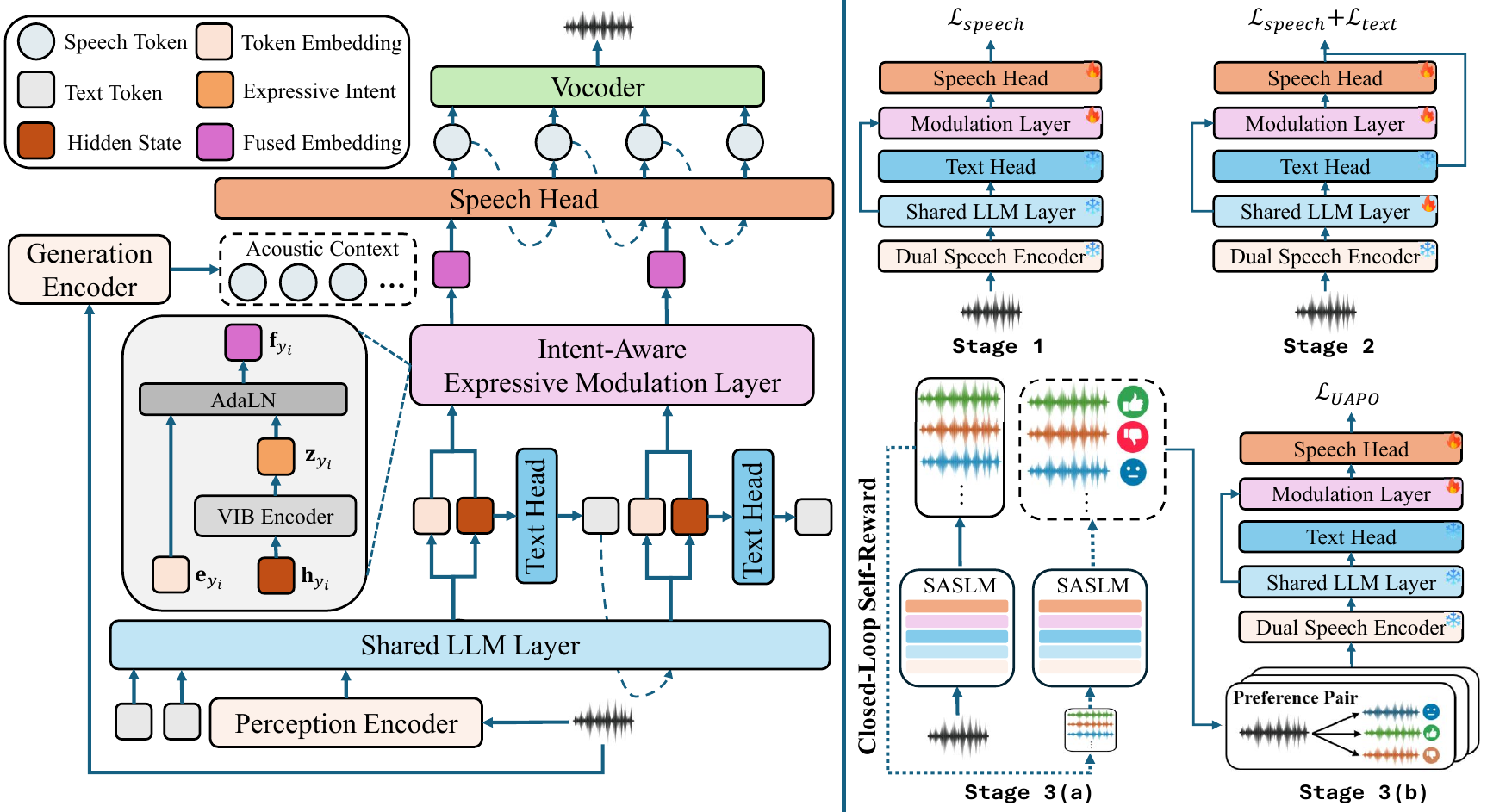}
\caption{
Overview of the SASLM framework. 
\textbf{Left:} Intent-aware architecture with VIB-driven expressive modulation. 
\textbf{Right:} Three-stage realization-aware training paradigm, progressing from acoustic bootstrapping to closed-loop self-reward optimization.
}
\label{fig:framework}
\end{figure*}

As shown in Fig.~\ref{fig:framework}, SASLM bridges semantic understanding and acoustic realization through two key designs: an intent-aware expressive modeling architecture (Sec.~\ref{sec:intent_aware_architecture}) and a realization-aware alignment training paradigm (Sec.~\ref{sec:3_stage_train}).

\subsection{Intent-Aware Expressive Modeling Architecture}
\label{sec:intent_aware_architecture}


As illustrated in the left part of Fig.~\ref{fig:framework}, SASLM follows the modular multi-head paradigm (Sec.~\ref{sec:related_work_slm}) and consists of three key components: (1) a semantic understanding backbone for multimodal perception and text response generation; (2) a VIB-based intent-aware expressive modulation layer that distills latent expressive intents $\mathbf{z}_{y_i}$ from contextual semantic hidden states $\mathbf{h}_{y_i}$ to modulate token embeddings $\mathbf{e}_{y_i}$, yielding intent-aware fused embeddings $\mathbf{f}_{y_i}$; and (3) the speech head generates intent-modulated speech by predicting speech tokens from the fused embeddings and acoustic context $\mathbf{R}_G$, yielding context-aware expressive speech.


\subsubsection{Semantic Understanding Backbone}
\label{sec:semantic_backbone}

\textbf{Semantic Perception and Understanding.}
Given a multimodal input $X$ comprising speech $X^{S}$ and/or text $X^{T}$, the model jointly generates a textual response $Y^{T}=\{y^{T}_1,\dots,y^{T}_N\}$ and a speech response $Y^{S}=\{y^{S}_1,\dots,y^{S}_M\}$. For the speech modality, a perception encoder $\phi_P$ extracts semantic representations $\mathbf{R}_P=\phi_P(X^S)$, which are fused with the text input to form a unified multimodal representation $\mathbf{E}_X$:
\begin{equation}
\mathbf{E}_X = \left[\mathrm{Embed}(X^T);\ \mathrm{Proj}(\mathbf{R}_P)\right],
\end{equation}
where $\mathrm{Embed}(\cdot)$ is the embedding lookup, $\mathrm{Proj}(\cdot)$ is linear projection and $[\cdot;\cdot]$ denotes concatenation.

The shared LLM backbone $\mathcal{H}$ then autoregressively models the multimodal context. At decoding step $i$, conditioned on $\mathbf{E}_X$ and the previously generated embeddings $\mathbf{e}_{y_{1:i}}$ with $\mathbf{e}_{y_i}=\mathrm{Embed}(y_i^T)$, it yields:
\begin{equation}
  \mathbf{h}_{y_i} = \mathcal{H}\!\left(\mathbf{E}_X,\,\mathbf{e}_{y_{1:i}}\right).
\end{equation}
These hidden states capture contextual semantics and serve as shared representations driving both text and speech generation.

\textbf{Text Generation.}
The text head decodes each hidden state $\mathbf{h}_{y_i}$ into a distribution over the text vocabulary to produce token $y^{T}_i$.

\subsubsection{Intent-aware Expressive Modulation}
\label{sec:vib_intent}
To let the model exploit its own semantic understanding for context-aware expressive realization, i.e., to be \textit{aware of what it thinks}, we introduce a VIB-based encoder $q_\phi$ that self-distills the LLM hidden states $\mathbf{H}_Y=[\mathbf{h}_{y_1},\dots,\mathbf{h}_{y_N}]$ into latent expressive intents $\mathbf{Z}_Y$, as shown in Fig.~\ref{fig:framework}. The bottleneck encourages $\mathbf{Z}_Y$ to  retain expressive cues such as emphasis and prosody, while discarding lexical information already captured by the token embeddings $\mathbf{E}_Y$. The distilled intent then modulates $\mathbf{E}_Y$ to produce intent-aware fused embeddings $\mathbf{F}_Y$, which drive speech generation together with the generation context $\mathbf{R}_G$ (Sec.~\ref{sec:intent_modulated_generation}). The ideal information-bottleneck objective is:
\begin{equation}
\scalebox{0.9}{$
\mathcal{J}
=
\max_{\theta}\;
I(\mathbf{F}_Y;\, Y^S \mid \mathbf{R}_{G}) %
-
\beta\, I(\mathbf{Z}_Y;\, \mathbf{E}_Y),
$}
\label{eq:ideal_obj}
\end{equation}
where $\beta$ balances expressive acoustic fidelity (first term) and compression that suppresses lexical leakage from $\mathbf{E}_Y$ into $\mathbf{Z}_Y$ (second term).

Directly optimizing Eq.~\ref{eq:ideal_obj} is intractable. Since the VIB encoder takes $\mathbf{H}_Y$ as input, $\mathbf{Z}_Y$ is conditionally independent of $\mathbf{E}_Y$ given $\mathbf{H}_Y$, forming the Markov chain $\mathbf{E}_Y \rightarrow \mathbf{H}_Y \rightarrow \mathbf{Z}_Y$. By the Data Processing Inequality (DPI)~\cite{cover1999elements}, $I(\mathbf{Z}_Y;\mathbf{E}_Y) \leq I(\mathbf{Z}_Y;\mathbf{H}_Y)$, so we penalize this tractable upper bound instead. Meanwhile, the realization term is lower-bounded by the conditional reconstruction loss of the speech head. Together, these yield the final variational surrogate objective (see Appendix~\ref{app:vib_bound_analysis} for derivation):
\begin{equation}
\scalebox{0.85}{$
\begin{aligned}
\tilde{\mathcal{J}}_{\mathrm{VIB}} = \max_{\theta,\phi}\; & \underbrace{ \mathbb{E}_{q_\phi} \left[ \log p_\theta ( Y^S \mid \mathbf{F}_Y, \mathbf{R}_{G} ) \right] }_{\text{expressive realization}} \\
& - \beta \underbrace{ D_{\mathrm{KL}} \left( q_\phi(\mathbf{Z}_Y \mid \mathbf{H}_Y) \,\Vert\, p(\mathbf{Z}_Y) \right) }_{\text{intent compression}}.
\end{aligned}
$}
\label{eq:vib_elbo}
\end{equation}

Since expressive intent unfolds smoothly over an utterance~\cite{nooteboom1997prosody, skerry2018towards}, we model $p(\mathbf{Z}_Y)$ with a discrete-time Ornstein--Uhlenbeck (OU) prior~\cite{uhlenbeck1930theory, li2020scalable}, which captures temporally correlated latent dynamics that the standard i.i.d. Gaussian prior cannot represent:
\begin{equation}
\scalebox{0.9}{$
p(\mathbf{z}_{y_i} \mid \mathbf{z}_{y_{i-1}})
=
\mathcal{N}\!\left(
\alpha \mathbf{z}_{y_{i-1}},
\sigma_p^2 \mathbf{I}
\right),
\quad \alpha \in (0,1),
$}
\label{eq:ou_prior}
\end{equation}
where $\alpha$ controls temporal persistence and $\sigma_p^2$ is the per-step transition variance.

Paired with this OU prior, we instantiate the VIB encoder by parameterizing the variational posterior as a token-wise diagonal Gaussian, i.e., $q_\phi(\mathbf{z}_{y_i} \mid \mathbf{h}_{y_i}) = \mathcal{N}(\boldsymbol{\mu}_i, \operatorname{diag}(\boldsymbol{\sigma}_i^2))$, whose mean and log-variance are predicted from the hidden state $\mathbf{h}_{y_i}$ by a lightweight MLP $f_\phi$. A differentiable intent is then drawn via the reparameterization trick:
\begin{equation}
\scalebox{0.75}{$
\mathbf{z}_{y_i}
=
\boldsymbol{\mu}_i
+
\boldsymbol{\sigma}_i \odot \boldsymbol{\epsilon}_i,
\quad
[\boldsymbol{\mu}_i,\boldsymbol{\sigma}_i]
=
f_{\phi}(\mathbf{h}_{y_i}),
\quad
\boldsymbol{\epsilon}_i \sim \mathcal{N}(\mathbf{0},\mathbf{I}),
$}
\label{eq:z_reparameterization}
\end{equation}
where $\odot$ denotes element-wise multiplication.

Given the OU prior (Eq.~\ref{eq:ou_prior}) and the Gaussian posterior (Eq.~\ref{eq:z_reparameterization}), the intent compression term (i.e., the KL divergence in Eq.~\ref{eq:vib_elbo}) can be computed analytically at each time step $i$ as:
\begin{equation}
\scalebox{0.8}{$
\mathcal{L}_{\mathrm{KL}}(i) = \frac{1}{2}\sum\limits_{k=1}^{d}\left[ \log \frac{\sigma_p^2}{\sigma_{i,k}^2} + \frac{\sigma_{i,k}^2 + (\mu_{i,k} - \alpha\,\mu_{i-1,k})^2}{\sigma_p^2} - 1 \right],
$}
\label{eq:kl_ou}
\end{equation}
where $\boldsymbol{\mu}_i, \boldsymbol{\sigma}_i \in \mathbb{R}^d$ are the per-dimension mean and standard deviation, with $\mu_{i,k}$ and $\sigma_{i,k}$ denoting their $k$-th components. This loss drives the expressive intent $\mathbf{z}_{y_i}$ to follow a temporally smooth distribution across timesteps. See Appendix~\ref{sec:appendix_vib} for the full derivation.

After extracting the latent intent $\mathbf{z}_{y_i}$ via the VIB encoder (Eq.~\ref{eq:z_reparameterization}), it is injected into the lexical token embeddings $\mathbf{e}_{y_i}$ through Adaptive Layer Normalization (AdaLN)~\cite{peebles2023scalable}:
\begin{equation}
\scalebox{0.75}{$\displaystyle
\mathbf{f}_{y_i} = \text{AdaLN}(\mathbf{e}_{y_i}, \mathbf{z}_{y_i}) = (1+\gamma(\mathbf{z}_{y_i})) \odot \frac{\mathbf{e}_{y_i}-\mu}{\sigma} + \delta(\mathbf{z}_{y_i})
$}
\end{equation}
where $\gamma(\cdot)$ and $\delta(\cdot)$ are zero-initialized linear projections, and $\mu, \sigma$ are the mean and standard deviation of $\mathbf{e}_{y_i}$.
This lightweight modulation preserves lexical content while injecting
token-level expressive guidance into speech conditioning.

\subsubsection{Intent-Modulated Speech Generation}
\label{sec:intent_modulated_generation}
The speech head autoregressively predicts speech tokens conditioned on the intent-modulated representations $\mathbf{f}_{y_i}$ and context $\mathbf{R}_{G}$:
\begin{equation}
\scalebox{0.75}{$\displaystyle
P_{\theta}(y^{S}_j \mid y^{S}_{<j},\, \mathbf{F}_{Y},\, \mathbf{R}_{G})
=
\text{SpeechHead}(y^{S}_{<j},\, \mathbf{F}_{Y},\, \mathbf{R}_{G}).
$}
\label{eq:speech_generation}
\end{equation}

We train the model by minimizing the negative variational lower bound in Eq.\ref{eq:vib_elbo}, which decomposes into an autoregressive speech reconstruction term and the VIB regularizer in Eq.\ref{eq:kl_ou}  that encourages temporally smooth expressive intent:
\begin{equation}
\scalebox{0.7}{$
\mathcal{L}_{\mathrm{speech}} = \underbrace{ -\sum_{t=1}^{M} \log p_{\theta} (y^S_t \mid \mathbf{F}_Y,\mathbf{R}_G,y^S_{<t}) }_{\mathcal{L}_{\mathrm{Recon}}} + \beta \underbrace{ \sum_{i=1}^{N} \mathcal{L}_{\mathrm{KL}}(i) }_{\mathcal{L}_{\mathrm{VIB}}}.
$}
\label{eq:speech_loss}
\end{equation}

To further adapt speech generation to the user's acoustic state, as the similar content with different paralinguistic cues may require distinct realizations~\cite{WynnB22},
we introduce a generation encoder $\phi_G$ that directly encodes raw input speech: $\mathbf{R}_{G} = \phi_G(X^S)$. Unlike the semantics-oriented perception pathway in Fig.~\ref{fig:diff_arch_slm}(a),
$\mathbf{R}_G$ preserves continuous paralinguistic cues, providing complementary
acoustic grounding for context-adaptive speech generation.

\subsection{Realization-Aware Alignment Training Paradigm}
\label{sec:3_stage_train}

We adopt a progressive realization-aware alignment training paradigm that gradually establishes the connection between semantic understanding and expressive speech generation.

\subsubsection{Stage 1: Acoustic Bootstrapping}
\label{sec:stage1_sft}
In this stage, the model establishes basic speech generation ability conditioned on latent intent representations and acoustic context. Only the modulation layer and speech head are trainable, learning to map intent-modulated semantic representations into speech tokens via the combined objective in Eq.~\ref{eq:speech_loss}. This builds the initial pathway from semantic understanding to acoustic realization.

\subsubsection{Stage 2: Expressive Intent Grounding}
\label{sec:stage2_expressive_sft}

Since the LLM backbone remains frozen in the first stage, its hidden states are primarily optimized for semantic modeling. We thus unfreeze its last layer to jointly optimize text prediction and speech reconstruction, grounding semantic representations in expressive speech generation without compromising the backbone's intelligence:
\begin{align}
\mathcal{L}_{\text{text}}
&=
-\sum_{i=1}^{N}
\log p_{\theta}(y^T_i \mid y^T_{<i},\mathbf{E}_X),
\\
\mathcal{L}_{\text{stage2}}
&=
\mathcal{L}_{\text{text}}
+
\mathcal{L}_{\text{speech}}.
\end{align}

This joint optimization progressively shifts LLM hidden states from semantic-dominant to intent-aware acoustically usable representations.

\subsubsection{Stage 3: Closed-Loop Self-Reward Alignment}
\label{sec:stage3_rl}

After the first two stages, training remains based on open-loop supervised reconstruction, lacking explicit cross-modal verification of whether generated speech faithfully realizes the expressive intent inferred from dialogue context. To address this, we introduce a closed-loop realization-aware alignment paradigm (details are provided in Algorithm~\ref{alg:self_refinement}, Appendix~\ref{app:algorithm}) comprising two steps:

\textbf{Stage 3(a): Self-Reward Generation.}
For each speech input $X^S$, SASLM generates a text response $Y^T$ together with multiple candidate speech rollouts $\{Y_k^S\}_{k=1}^{K}$. A TTS system additionally produces a context-unaware anchor response, as a minimum acceptable prospect~\cite{Ethayarajh2024KTOMA} baseline $Y^S_\perp$. The SLM then evaluates each rollout using rubric-based criteria covering emotion, prosody, and naturalness, while WER is used as a hard intelligibility constraint~\cite{zhang2025speechjudge}. Rollouts are then partitioned into the preferred set $\mathcal{Y}_w$ and the rejected set $\mathcal{Y}_l$  for subsequent alignment according to their scores relative to the anchor $Y^S_\perp$.

\textbf{Stage 3(b): Preference-based Alignment.}
We further use the obtained preference pairs to align expressive speech generation via Utility-Anchored Preference Optimization (UAPO)~\cite{wang2025adaptive}, which promotes preferred expressive realizations over context-unaware baseline generations and suppresses rejected responses:
\begin{equation}
\begin{gathered}
\scalebox{0.85}{$\displaystyle
u_\theta(x, y) = \frac{1}{|y|} \sum_{i=1}^{|y|} 
    \log \pi_\theta(y_i \mid x,\, y_{<i})$} \\[4pt]
\scalebox{0.65}{$\displaystyle
\mathcal{L}_w 
= -\sum_{y_w \in \mathcal{Y}_w} 
\log \frac{
    \exp\bigl(u_\theta(X, y_w)\bigr)
}{
    \exp\bigl(u_\theta(X, Y^S_\perp)\bigr) 
    + \sum_{y' \in \mathcal{Y}_w} \exp\bigl(u_\theta(X, y')\bigr)
}$} \\[4pt]
\scalebox{0.75}{$\displaystyle
\mathcal{L}_l 
= -\log \frac{
    \exp\bigl(u_\theta(X,\, Y^S_\perp)\bigr)
}{
    \exp\bigl(u_\theta(X,\, Y^S_\perp)\bigr) 
    + \sum_{y_l \in \mathcal{Y}_l} \exp\bigl(u_\theta(X,\, y_l)\bigr)
}$} \\[4pt]
\scalebox{0.85}{$\displaystyle
\mathcal{L}_{\text{UAPO}} 
= \mathbb{E}_{(X,\,\mathcal{Y}_w,\,\mathcal{Y}_l,\,Y^S_\perp) \sim \mathcal{B}}
\bigl[\mathcal{L}_w + \mathcal{L}_l\bigr].
$}
\end{gathered}
\label{eq:loss_uapo}
\end{equation}

\section{Experiments}


\subsection{Experimental Setups}

\paragraph{Model Configuration.}

\begin{table}[t]
\centering
\resizebox{\columnwidth}{!}{%
\begin{tabular}{lccc}
\toprule
Dataset & Stage & \#Samples & Hours \\
\midrule
InstructS2S~\cite{fang2025llamaomni} & I   & 200k & $\sim$2000 h \\
Genshin Dataset~\footnotemark              & II  & 220k & $\sim$603 h \\
EmoVoice-DB~\cite{yang2025emovoice} & II  & 66k  & $\sim$192 h  \\
EmoNet~\cite{schuhmann2025emonet}   & III & 200k & $\sim$263 h \\
\bottomrule
\end{tabular}%
}
\caption{Datasets used in different training stages.}
\label{tab:dataset}
\end{table}
\footnotetext{\url{https://www.modelscope.cn/datasets/aihobbyist/Genshin_Dataset}}

We adopt the Thinker module of Qwen2.5-Omni-3B~\cite{xu2025qwen25omnitechnicalreport} as the semantic understanding backbone, which employs Whisper as the perception encoder and Qwen2.5-3B as the shared LLM layers with its text head. The speech head is initialized from CosyVoice2~\cite{du2024cosyvoice}, reusing its S3 tokenizer as the generation encoder and its flow matching module and vocoder for waveform synthesis. Stage-specific datasets are used to support progressive realization-aware alignment, as summarized in Table~\ref{tab:dataset}. Dataset construction and training configurations are detailed in Appendix~\ref{app:appendix_implementation}.

\paragraph{Evaluation Datasets.}
We primarily evaluate expressive speech generation on EchoMind~\cite{zhou2025echomind}, an empathy-oriented spoken dialogue benchmark covering seven emotion categories. To verify that expressive generation does not degrade semantic understanding, we further evaluate on the multimodal audio benchmark MMAU~\cite{sakshi2024mmau} and assess empathetic response quality on EchoMind provided in Appendix~\ref{sec:appendix_exp_of_semantic_understanding}.

\paragraph{Baselines.}
(1) \textbf{Controllable TTS}, representing oracle expressive control, including CosyVoice2~\cite{du2024cosyvoice} and GPT-4o-mini-TTS~\cite{hurst2024gpt}, which synthesize speech from reference replies using predefined TTS prompts; (2) \textbf{Vanilla SLMs}, performing direct semantic-to-speech generation without explicit expressive modeling, including Qwen2.5-Omni (3B/7B)~\cite{xu2025qwen25omnitechnicalreport}, Qwen3-Omni-30B~\cite{xu2025qwen3omnitechnicalreport}, GLM-4-Voice~\cite{zeng2024glm}, Step-Audio2-Mini~\cite{huang2025step}, Kimi-Audio~\cite{ding2025kimi}, and UltraVoice-7B~\cite{tu2025ultravoice}, which is trained on style-controlled expressive data; (3) \textbf{Emotional SLMs}, adopting proxy-based generation, represented by EMOVA-7B~\cite{chen2025emova}, which produces emotion prompts prior to synthesis; (4) \textbf{Commercial SLMs}, represented by Doubao~\footnote{https://www.volcengine.com/docs/6561/1594356} as an industrial reference.

\subsection{Evaluation Protocol}
\paragraph{Utterance-level Objective Evaluation.}
To evaluate expressive acoustic realization, we report four categories of objective metrics:
(1) \textbf{Intelligibility}, measured by Word Error Rate (WER);
(2) \textbf{Expressiveness}, measured by pitch variance (F0-Var)~\cite{busso2009analysis};
(3) \textbf{Semantic–acoustic emotion alignment}, measured by EmoAlign with emotion2vec~\cite{ma2024emotion2vec} to assess the consistency between text and speech emotions in the response;
and (4) \textbf{Speech naturalness}, evaluated by Audiobox-Aesthetics~\cite{tjandra2025meta}, including Content Enjoyment (CE), Usefulness (CU), Production Complexity (PC), Quality (PQ), and their average score (Aes.).

\paragraph{Conversation-level Subjective Evaluation.}
Following~\cite{tu2025ultravoice}, we employ Gemini-2.5-Flash~\cite{comanici2025gemini} as an SLM-as-Judge on a 1–5 scale to evaluate whether the generated speech exhibits contextually appropriate emotion (Emo.), prosody (Pro.), naturalness (Nat.), and overall interaction quality (Ovr.). Prompt details are provided in Appendix~\ref{app:appendix_prompt}.

\subsection{Main Results}
\label{sec:main_result}

\begin{table*}[htbp]
\centering
\small
\resizebox{\textwidth}{!}{%
\begin{tabular}{llccccccccc}
\toprule
\multirow{2}{*}{\textbf{Model Type}} 
& \multirow{2}{*}{\textbf{Model Name}} 
& \multicolumn{4}{c}{\textbf{Utterance-Level Objective Metrics}} 
& \multicolumn{4}{c}{\textbf{Conversation-Level Subjective Metrics}} \\
\cmidrule(lr){3-6} \cmidrule(lr){7-10}
& 
& \textbf{WER\%}~$\downarrow$ 
& \textbf{F0-Var}~$\uparrow$ 
& \textbf{EmoAlign\%}~$\uparrow$ 
& \textbf{CE / CU / PC / PQ / Aes.}~$\uparrow$ 
& \textbf{Emo.}~$\uparrow$ 
& \textbf{Pro.}~$\uparrow$ 
& \textbf{Nat.}~$\uparrow$ 
& \textbf{Ovr.}~$\uparrow$ \\
\midrule
\multirow{2}{*}{\makecell[l]{Controllable TTS \\ (w/ Oracle Prompts)}}
& CosyVoice2
& 3.80 & 51.20 & 39.07
& 6.21 / 7.09 / 1.42 / 7.67 / 5.60 
& 4.15 & 4.25 & 4.34 & 4.23 \\
& GPT-4o-mini-TTS
& 7.63 & 70.44 & 45.04
& 6.22 / 7.37 / 1.41 / 8.00 / 5.75 
& 4.27 & 4.43 & 4.58 & 4.41 \\
\midrule
Commercial SLM 
& Doubao
& 1.12 & 68.58 & 42.95
& 6.21 / 7.38 / 1.53 / 7.94 / 5.72
& 4.22 & 4.50 & 4.59 & 4.44 \\
\midrule
\multirow{7}{*}{Vanilla SLM}
& Kimi Audio 
& 6.99 & 46.98 & 22.64
& 5.31 / 5.98 / \textbf{1.49} / 6.87 / 4.91 
& 3.77 & 4.02 & 4.15 & 3.97 \\
& Step Audio2 Mini 
& 4.87 & \underline{50.92} & 22.92
& 6.12 / 7.06 / 1.44 / 7.68 / 5.58 
& 4.00 & 4.19 & \underline{4.43} & 4.15 \\
& GLM-4-Voice 
& 8.88 & 43.71 & 23.84
& \underline{6.22} / 6.94 / 1.42 / 7.57 / 5.54 
& 3.95 & 4.11 & 4.29 & 4.09 \\
& UltraVoice-7B 
& 4.21 & 49.53 & 17.34
& 6.11 / 7.13 / 1.44 / 7.55 / 5.56 
& 4.01 & 4.17 & 4.26 & 4.11 \\
& Qwen2.5-Omni-3B 
& 4.14 & 42.88 & 21.44
& 6.15 / 6.90 / \underline{1.47} / 7.74 / 5.57 
& 3.71 & 3.84 & 3.96 & 3.79 \\
& Qwen2.5-Omni-7B 
& \underline{3.75} & 47.77 & 23.10
& \textbf{6.23} / 7.08 / 1.42 / \underline{7.79} / \underline{5.63} 
& 3.62 & 3.90 & 4.09 & 3.86 \\
& Qwen3-Omni-30B 
& \textbf{2.76} & 49.76 & 25.03
& \textbf{6.23} / \underline{7.14} / 1.41 / 7.70 / 5.62 
& 4.08 & \underline{4.28} & 4.42 & \underline{4.25} \\
\midrule
Emotional SLM
& EMOVA-7B
& 10.09 & 47.98 & \underline{30.58}
& 5.68 / 6.45 / \textbf{1.49} / 6.84 / 5.12
& \underline{4.10} & 4.01 & 4.14 & 4.06 \\
\midrule

\rowcolor{gray!15} 
Expressive SLM
& SASLM-3B (Ours)
& 4.56 & \textbf{63.44} & \textbf{35.61}
& 6.12 / \textbf{7.19} / 1.41 / \textbf{7.82} / \textbf{5.64} 
& \textbf{4.24} & \textbf{4.36} & \textbf{4.49} & \textbf{4.33} \\
\bottomrule
\end{tabular}%
}
\caption{%
Comparison of context-aware expressive speech generation.
\textbf{Bold} and \underline{underline} denote the best and second-best results among \textbf{open-source SLMs}, excluding commercial SLMs and controllable TTS systems.
}
\label{tab:slm_comparison}
\end{table*}

As shown in Table~\ref{tab:slm_comparison}, SASLM achieves the best overall performance among open-source SLMs and rivals commercial systems such as GPT-4o-mini-TTS and Doubao, a finding further corroborated by human evaluation in Appendix~\ref{app:human_evaluation}. With only 3B parameters, SASLM consistently surpasses much larger vanilla SLMs, including Qwen3-Omni-30B, delivering richer pitch dynamics (F0-Var: 63.44 vs.\ 49.76), stronger semantic–acoustic emotion alignment (EmoAlign: 35.61\% vs.\ 25.03\%), and higher subjective overall quality (4.33 vs.\ 4.25). While model scaling and improved semantic intelligence enhance content fidelity and robustness (e.g., Qwen3-Omni-30B's lower WER of 2.76), they do not automatically translate into expressive acoustic realization. This suggests that explicit modeling of expressive intent is more effective than relying solely on implicit semantic learning at scale.

Compared with Controllable TTS systems using oracle expressive prompts as an idealized alignment baseline, SASLM achieves higher subjective prosody (4.36 vs. 4.25), naturalness (4.49 vs. 4.34), and overall interaction quality (4.33 vs. 4.23), despite CosyVoice2 showing slightly better objective emotion alignment with ground-truth emotion labels (39.07\% vs. 35.61\%). This suggests that oracle labels mainly impose coarse-grained emotion consistency, while SASLM dynamically infers token-level, context-aware expressive intent for finer speech realization, enabling end-to-end SLMs to approach or even surpass oracle-prompted TTS in subjective quality. The similar trend between GPT-4o-mini-TTS and Doubao further supports the benefit of stronger end-to-end expressive modeling for context-aware speech generation.

SASLM also outperforms emotional SLMs that rely on external expressive proxies. Compared with EMOVA-7B, which conditions on self-generated emotion labels, SASLM achieves notable gains in prosody (4.36 vs. 4.01), overall quality (4.33 vs. 4.06), and emotional intent alignment (EmoAlign: 35.61\% vs. 30.58\%). These results indicate that intrinsic latent intent provides  richer, more fine-grained expressive guidance  than discrete emotion-centric discrete proxies under explicit supervision.

\section{Analysis}
\begin{figure}[t]
    \centering
    \includegraphics[width=0.45\textwidth]{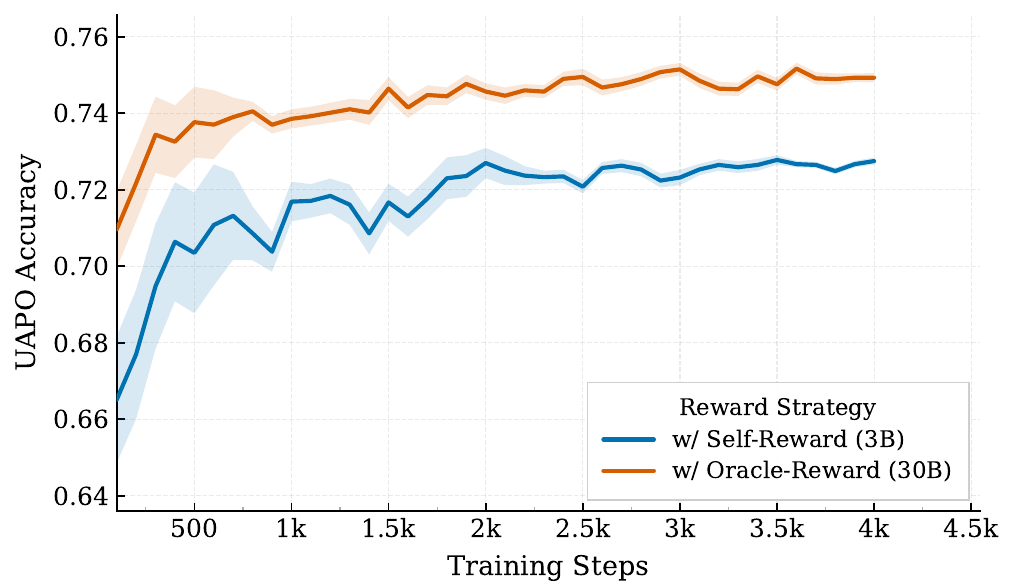}
    \caption{Evaluation accuracy dynamics during UAPO training with different reward models.}
    \label{fig:rl_accuracy_curve}
\end{figure}

\begin{figure*}[t]
    \centering
    \includegraphics[width=\textwidth]{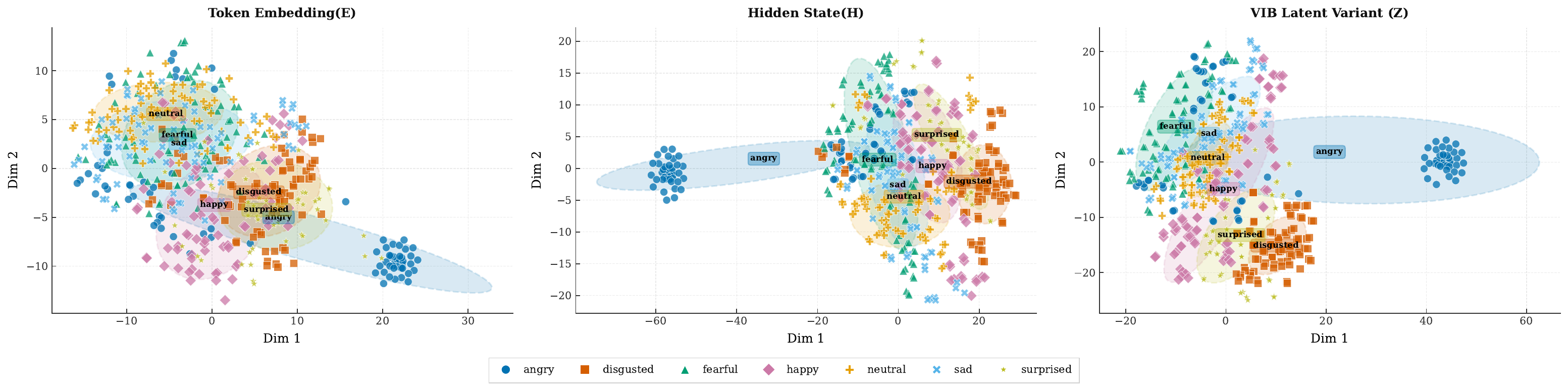}
    \caption{
t-SNE visualization of emotion clustering across token embeddings ($\mathbf{e}$), LLM hidden states ($\mathbf{h}$), and distilled VIB latent variables ($\mathbf{z}$).
Linear-probe accuracies, computed on the original representations rather than the t-SNE projections, are 18.1\%, 41.4\%, and 64.5\%, respectively.
}
    \label{fig:analysis_emotion_cluster}
\end{figure*}

\subsection{Architectural View: Intent Modulation Meets Acoustic Grounding}
\label{sec:ablation_study_architecture}

Table~\ref{tab:ablation_fusion} ablates two architectural factors after the first two training stages: 
(1) the \textbf{intent modulation strategy} $\mathcal{M}(\mathbf{e}_{y_i}, \mathbf{h}_{y_i})$, for which we compare a progression of increasingly expressive designs: a conversational TTS-style baseline $\pi_1$ using only content embeddings $\mathbf{e}_{y_i}$, naive element-wise fusion $\oplus$~\cite{xu2025qwen25omnitechnicalreport}, adaptive modulation via AdaLN, and our proposed VIB+AdaLN via a self-distill-then-modulate scheme
and (2) the \textbf{context-grounding representation} $\mathbf{R}_G$, where $\varnothing$, \textit{Semantic}, and \textit{Acoustic} denote no context, perception-pathway representations ($\mathbf{R}_G=\mathbf{E}_X$), and generation-encoder acoustic features ($\mathbf{R}_G=\phi_G(X^S)$), respectively.

For intent modulation, directly adding semantic hidden states yields marginal gains over $\pi_1$ (Ovr.\: 4.20 vs.\ 4.17), but increases WER from
3.61\% to 8.29\%. AdaLN mitigates this instability (WER 6.50\%), whereas the
proposed VIB-based AdaLN achieves the best overall quality (Ovr.: 4.31) with
a lower WER of 4.74\%. This suggests that VIB improves the
expression–fidelity trade-off by disentangling expressive intent from
lexical content for more stable expressive speech generation.

For context grounding, removing context yields a cleaner but more conservative setting, slightly improving WER (4.19\%) but reducing context-appropriate overall quality (4.21). Acoustic grounding improves the overall score to 4.31 with only a modest WER increase (4.74\%), underscoring its role in context-adaptive expressive realization. In contrast, semantic grounding severely harms intelligibility (WER 11.97\%) without improving overall quality (4.20), suggesting that injecting semantic features alone perturbs acoustic modeling and destabilizes expressive realization. Full results are provided in Appendix~\ref{sec:appendix_complete_ablation_study}.

\subsection{Optimization View: Efficient Alignment via Closed-Loop Self-Reward}
\begin{table}[t]
\centering
\resizebox{\columnwidth}{!}{
\begin{tabular}{lll cccc}
\toprule
$\mathcal{M}(\mathbf{e}_{y_i}, \mathbf{h}_{y_i})$
  & $\mathbf{R}_{G}$
  & \textbf{WER\%$\downarrow$}
  & \textbf{Emo.$\uparrow$}
  & \textbf{Pro.$\uparrow$}
  & \textbf{Nat.$\uparrow$}
  & \textbf{Ovr.$\uparrow$} \\
\midrule
\multicolumn{7}{l}{\textit{(A) Ablation on Modulation Strategy}} \\
\midrule
$\pi_1$   & Acoustic & \textbf{3.61} & 4.06 & 4.22 & 4.30 & 4.17 \\
$\oplus$  & Acoustic & 8.29 & 4.04 & \underline{4.28} & 4.31 & 4.20 \\
$\mathrm{AdaLN}$     & Acoustic & 6.50 & \underline{4.15} & \underline{4.28} & 4.34 & \underline{4.22} \\

\midrule
\multicolumn{7}{l}{\textit{(B) Ablation on Context Grounding}} \\
\midrule

$\mathrm{VIB+AdaLN}$ & $\varnothing$  & \underline{4.19} & 4.10 & 4.23 & 4.33 & 4.21 \\
$\mathrm{VIB+AdaLN}$ & Semantic & 11.97 & 4.08 & 4.24 & \underline{4.36} & 4.20 \\

\midrule
\rowcolor{gray!15}
$\mathrm{VIB+AdaLN}$         & Acoustic & 4.74
                             & \textbf{4.21} & \textbf{4.34} 
                             & \textbf{4.45} & \textbf{4.31} \\
\bottomrule
\end{tabular}}
\caption{
Ablation study on the modulation strategy $\mathcal{M}(\mathbf{e}_{y_i}, \mathbf{h}_{y_i})$ and context grounding $\mathbf{R}_G$ in Eq.~\ref{eq:speech_generation}.
}
\label{tab:ablation_fusion}
\end{table}
\begin{table}[ht]
\centering
\small
\resizebox{\columnwidth}{!}{%
\begin{tabular}{lc c cccc}
\toprule
\textbf{Strategy}
  & \textbf{RM Size}
  & \textbf{WER(\%)$\downarrow$}
  & \textbf{Emo.$\uparrow$}
  & \textbf{Pro.$\uparrow$}
  & \textbf{Nat.$\uparrow$}
  & \textbf{Ovr.$\uparrow$} \\
\midrule
SA-SLM(SFT)    & ---  & 4.74          & 4.21          & 4.34          & 4.45          & 4.31 \\
\quad + Self-Reward     & 3B   & 4.56          & 4.24          & 4.36          & 4.49          & 4.33 \\
\quad + Oracle-Reward   & 30B  & \textbf{4.21} & \textbf{4.30} & \textbf{4.41} & \textbf{4.54} & \textbf{4.38} \\
\bottomrule
\end{tabular}
}
\caption{Comparison of reward strategies on Intelligibility (WER) and expressive speech quality.}
\label{tab:reward_model}
\end{table}

As shown in Table~\ref{tab:reward_model}, realization-aware alignment with both the 3B Self-Reward and 30B Oracle-Reward scorers consistently outperforms the open-loop SFT baseline, confirming its effectiveness. Notably, expressive quality and intelligibility improve simultaneously during alignment (Ovr.\ 4.31 $\rightarrow$ 4.38; WER 4.74\% $\rightarrow$ 4.21\%), suggesting that closed-loop optimization mitigates the expressivity--stability trade-off and scales reliably.

As shown in Fig.~\ref{fig:rl_accuracy_curve}, the 30B critic converges faster and more stably, indicating stronger discrimination of high-quality rollouts and reduced preference-label noise, consistent with the human preference-accuracy results in Appendix~\ref{app:reward_validation}. Notably, the 3B critic also improves steadily, showing that smaller models can still provide effective alignment signals through the self-reward mechanism, enabling understanding-guided expressive realization without compromising intelligibility.

\subsection{Representational View: How VIB Distills Expressive Intent}

To examine how VIB reshapes expressive representations, Fig.~\ref{fig:analysis_emotion_cluster} compares emotion separability across token embeddings $\mathbf{e}$, hidden states $\mathbf{h}$, and VIB latent intents $\mathbf{z}$. Token embeddings show heavily overlapping clusters with near-random linear-probe accuracy (18.1\%), while hidden states improve separability (41.4\%), indicating that LLM semantic understanding contains expressive cues but remain entangled with noise. Notably, VIB latents achieve the clearest clustering and highest accuracy (64.5\%), showing that the OU-constrained VIB distills more discriminative expressive intent from entangled semantic states, yielding representations better aligned with expressive acoustic realization.

\section{Conclusion}
\label{sec:conclusion}
In this paper, we identify the gap between semantic understanding and acoustic realization in current SLMs and propose SASLM to bridge it via self-aware intent–realization alignment. With only 3B parameters and 800 hours of expressive data, SASLM achieves SOTA among open-source SLMs, surpassing the 10$\times$ larger Qwen3-Omni-30B in F0-Var (63.44 vs 49.76) and overall score (4.33 vs 4.25). Closed-loop alignment further boosts quality (4.31 $\rightarrow$ 4.38) while reducing WER (4.74\% $\rightarrow$ 4.21\%), easing the expressivity–stability trade-off. SASLM also outperforms proxy-based models like EMOVA-7B and rivals commercial SLM such as Doubao, showing that intent modeling efficiently enables expressive generation.


\section*{Limitations}
Although SASLM improves expressive speech generation without external emotion labels, several limitations remain. First, our evaluation mainly focuses on English empathetic dialogues in EchoMind, leaving multilingual, cross-cultural, and domain-specific expressive behaviors underexplored. Second, stronger expressive synthesis may make synthetic voices more persuasive, increasing risks of impersonation, emotional manipulation, and deceptive audio. Deployments should therefore incorporate synthetic-speech disclosure, watermarking or traceability when feasible, and safeguards for emotionally sensitive use cases.

\section*{Ethics Statement}
AI-assisted tools are used only for writing polishing and language refinement. All scientific contributions, experiments, conclusions, and citations are manually developed and verified by the authors. All training and evaluation data are from publicly available or open-source datasets. We strictly follow the original licenses, usage restrictions, and ethical requirements specified by the dataset providers, without applying additional special processing such as extra personal-information filtering.

\bibliography{custom}

\clearpage
\appendix

\section{Detailed Algorithm for Self-Reward Alignment}
\label{app:algorithm}

In this section, we provide the detailed pseudo-code for the closed-loop self-reward alignment paradigm introduced in Stage 3. Algorithm~\ref{alg:self_refinement} outlines the complete pipeline, including data construction, rubric-based evaluation, and preference pair selection. Specifically, we adopt the English subset of EmoNet~\cite{schuhmann2025emonet} as the question source. Given each question, SASLM first specifies the intended expression along three dimensions—emotional intensity, prosody, and naturalness. The generated rollouts are then scored against predefined rubrics to assess their adherence to the intended expression. The scoring template for the emotion dimension is illustrated in Figure~\ref{fig:emotion_rl_prompt}.

\begin{algorithm}[tb]
  \caption{Closed-Loop Self-Reward Alignment }
  \label{alg:self_refinement}
  \small \selectfont
  \begin{algorithmic}[1]
    \STATE {\bfseries Input:} Dataset $\mathcal{D}$, WER threshold $\tau$, 
    TTS model $\theta_{\text{TTS}}$, SLM $\theta_{\text{SLM}}$,
    prompts $p_{\text{emo}}$, $p_{\text{pro}}$, $p_{\text{nat}}$, $p_{\text{asr}}$,
    sample size $N$
    \STATE Initialize replay buffer $\mathcal{B} \leftarrow \emptyset$

    \STATE {\color{gray}\textit{\# Stage 3(a): Self-Reward}}
    \FOR{each $X^S \in \mathcal{D}$}
        \STATE {\color{gray}\textit{\# Generate the text response and $N$ speech rollouts}}
        \STATE $Y^T,\, \{Y^{S}_k\}_{k=1}^{N} \leftarrow \theta_{\text{SLM}}(X^S)$ 
        \STATE $Y^S_\perp \leftarrow \theta_{\text{TTS}}(Y^T)$ 
               \hfill{\color{gray}\textit{// Context-unaware anchor}}
        
        \STATE {\color{gray}\textit{\# Generate rubrics for each dimension}}
        \STATE $\mathcal{R}_d \leftarrow \theta_{\text{SLM}}(X^S,\, Y^T,\, p_d),
               \quad d \in \{\text{emo},\, \text{pro},\, \text{nat}\}$
               
        \STATE $\mathcal{R} \leftarrow \{\mathcal{R}_{\text{emo}},\, 
               \mathcal{R}_{\text{pro}},\, \mathcal{R}_{\text{nat}}\}$

        \STATE {\color{gray}\textit{\# Evaluate rollouts and anchor against rubrics}}
        \FOR{each $Y \in \{Y^{S}_k\}_{k=1}^{N} \cup \{Y^S_\perp\}$}
            \STATE $s_{\text{emo}},\, s_{\text{pro}},\, s_{\text{nat}} \leftarrow 
                   \theta_{\text{SLM}}(Y,\ \mathcal{R}) \in \{0,\, 0.5,\, 1\}^3$
            \STATE $\text{wer} \leftarrow \text{WER}\!\left(\theta_{\text{SLM}}
                   (Y,\, p_{\text{asr}}),\ Y^T\right)$
            \STATE $r(Y) \leftarrow 0$ \textbf{if} $\text{wer} > \tau$ 
                   \textbf{else} $(s_{\text{emo}} + s_{\text{pro}} + s_{\text{nat}})$
        \ENDFOR

        \STATE $\mathcal{G}_{+} \leftarrow \{ Y^{S}_k \mid r(Y^{S}_k) > r(Y^S_\perp) \}$
        \STATE $\mathcal{G}_{-} \leftarrow \{ Y^{S}_k \mid r(Y^{S}_k) < r(Y^S_\perp) \}$

        \IF{$\mathcal{G}_{+} \neq \emptyset$ \AND $\mathcal{G}_{-} \neq \emptyset$}
            \STATE $\mathcal{Y}_w \leftarrow \bigl\{ y \in \mathcal{G}_{+} \mid 
                   r(y) = \max_{y' \in \mathcal{G}_{+}} r(y') \bigr\}$
            \STATE $\mathcal{Y}_l \leftarrow \bigl\{ y \in \mathcal{G}_{-} \mid 
                   r(y) = \min_{y' \in \mathcal{G}_{-}} r(y') \bigr\}$
            \STATE $\mathcal{B} \leftarrow \mathcal{B} \cup 
                   \{(X^S,\ \mathcal{Y}_w,\ \mathcal{Y}_l,\ Y^S_\perp)\}$
        \ENDIF
    \ENDFOR

    \STATE {\color{gray}\textit{\# Stage 3(b): Offline Optimization}}
    \FOR{each $(X^S,\ \mathcal{Y}_w,\ \mathcal{Y}_l,\ Y^S_\perp) \in \mathcal{B}$}
        \STATE Compute $\mathcal{L}_{\text{UAPO}}$ via 
               Eq.~\eqref{eq:loss_uapo} and update $\theta_{\text{SLM}}$
    \ENDFOR

  \end{algorithmic}
\end{algorithm}

\section{Derivation of the VIB-Driven Disentanglement Objective}
\label{sec:appendix_vib}

Section~\ref{sec:vib_intent} introduced the ideal information bottleneck objective (Eq.~\ref{eq:ideal_obj}) and invoked the Data Processing Inequality (DPI)~\cite{cover1999elements} to derive a tractable surrogate (Eq.~\ref{eq:vib_elbo}). This appendix presents the complete derivation, including the Markov chain formulation, the variational bounds, and the closed-form KL divergence under an Ornstein--Uhlenbeck (OU)~\cite{uhlenbeck1930theory} prior.

\subsection{The Ideal Objective and the DPI Surrogate}

The intent-aware architecture aims to extract a pure expressive intent $\mathbf{Z}_Y$ that maximizes acoustic realization fidelity ($Y^S$) while discarding lexical redundancy already encoded in the text embeddings ($\mathbf{E}_Y$). This yields the ideal information bottleneck objective:

\begin{equation}
\scalebox{0.90}{$
\mathcal{J}
=
\max_{\theta}
\;
I(\mathbf{F}_Y;\, Y^S \mid \mathbf{R}_{G})
-
\beta\,
I(\mathbf{Z}_Y;\, \mathbf{E}_Y)
$}
\end{equation}

where $\mathbf{F}_Y$ is the intent-modulated representation, $\mathbf{R}_G$ the acoustic context, and $\beta$ the compression strength.

Since $I(\mathbf{Z}_Y;\, \mathbf{E}_Y)$ is intractable, we exploit the fact that $\mathbf{Z}_Y$ is sampled solely from the LLM hidden states $\mathbf{H}_Y$ via the VIB encoder $q_\phi(\mathbf{Z}_Y \mid \mathbf{H}_Y)$, with no direct access to $\mathbf{E}_Y$. The resulting conditional independence $p(\mathbf{Z}_Y \mid \mathbf{H}_Y, \mathbf{E}_Y) = p(\mathbf{Z}_Y \mid \mathbf{H}_Y)$ defines the Markov chain:
\begin{equation}
    \mathbf{E}_Y \to \mathbf{H}_Y \to \mathbf{Z}_Y.
\end{equation}
By the DPI, information cannot increase along a Markov chain, which guarantees that:
\begin{equation}
    I(\mathbf{Z}_Y;\, \mathbf{E}_Y) \leq I(\mathbf{Z}_Y;\, \mathbf{H}_Y),
\end{equation}
so $I(\mathbf{Z}_Y;\, \mathbf{H}_Y)$ serves as a tractable upper bound for penalization.

\paragraph{Physical Interpretation of the Surrogate.}
Substituting this upper bound into the objective yields the surrogate:

\begin{equation}
\scalebox{0.85}{$
\tilde{\mathcal{J}}
=
\max_{\theta,\phi}
\;
I(\mathbf{F}_Y;\,Y^S\mid\mathbf{R}_{G})
-
\beta\,
I(\mathbf{Z}_Y;\,\mathbf{H}_Y)
$}
\end{equation}

Penalizing $I(\mathbf{Z}_Y;\, \mathbf{H}_Y)$ rather than $I(\mathbf{Z}_Y;\, \mathbf{E}_Y)$ is in fact desirable: it compels $\mathbf{Z}_Y$ to discard not only redundant lexical tokens (already captured in $\mathbf{E}_Y$) but also irrelevant contextual semantics, retaining \textit{only} the expressive cues essential for acoustic realization.

\subsection{Variational Bounds for the Surrogate Objective}
\label{app:vib_bound_analysis}
As exact mutual information is intractable for high-dimensional neural representations, we derive variational bounds for both terms in $\tilde{\mathcal{J}}$.

\paragraph{Lower Bound for Expressive Realization.}
By applying the factorization
\[
p(\mathbf{F}_Y, Y^S \mid \mathbf{R}_{G}) = p(Y^S \mid \mathbf{F}_Y, \mathbf{R}_{G})\, p(\mathbf{F}_Y \mid \mathbf{R}_{G}),
\]
the conditional mutual information admits the decomposition:
\begin{equation}
\scalebox{0.65}{$
I(\mathbf{F}_Y;\, Y^S \mid \mathbf{R}_{G})
=
\mathbb{E}_{p}\!\left[
\log p(Y^S\mid\mathbf{F}_Y,\mathbf{R}_{G})
\right]
+
H(Y^S\mid\mathbf{R}_{G})
$}
\label{eq:cmi_split}
\end{equation}

where $\mathbb{E}_p$ denotes the expectation over the empirical data distribution. Since the true posterior $p(Y^S\mid\mathbf{F}_Y,\mathbf{R}_{G})$ is unknown, we introduce a parameterized decoder $p_\theta(Y^S\mid\mathbf{F}_Y,\mathbf{R}_{G})$ (i.e., the speech generation head). By the non-negativity of the KL divergence, $D_{\mathrm{KL}}(p\,\|\,p_\theta)\geq 0$, we directly obtain:

\begin{equation}
\scalebox{0.75}{$
\mathbb{E}_{p}\!\left[\log p(Y^S\mid\mathbf{F}_Y,\mathbf{R}_{G})\right]
\geq
\mathbb{E}_{p}\!\left[\log p_\theta(Y^S\mid\mathbf{F}_Y,\mathbf{R}_{G})\right]
$}
\label{eq:kl_bound}
\end{equation}

Substituting this into Eq.~\eqref{eq:cmi_split} yields the Variational Lower Bound (VLB). Since $H(Y^S\mid\mathbf{R}_{G})$ is independent of $\theta$, maximizing this bound is equivalent to minimizing the negative log-likelihood. Applying autoregressive factorization over the length-$M$ speech sequence yields the reconstruction loss:

\begin{equation}
\scalebox{0.90}{$
\mathcal{L}_{\mathrm{Recon}}
=
-\sum_{t=1}^{M}
\log p_\theta\!\left(
y_t^{S}
\mid
\mathbf{F}_Y,
\mathbf{R}_{G},
y_{<t}^{S}
\right)
$}
\label{eq:recon}
\end{equation}

Note that $\mathbf{R}_{G} = \phi_G(X^S)$, produced by an independent perception encoder, lies outside the Markov chain $\mathbf{E}_Y \!\to\! \mathbf{H}_Y \!\to\! \mathbf{Z}_Y$, leaving the DPI argument unaffected.

\paragraph{Upper Bound for Intent Compression.}
For the second term $I(\mathbf{Z}_Y;\, \mathbf{H}_Y)$, the mutual information is defined as:

\begin{equation}
\scalebox{0.90}{$
I(\mathbf{Z}_Y;\,\mathbf{H}_Y)
=
\mathbb{E}_{p(\mathbf{Z}_Y,\mathbf{H}_Y)}
\!\left[
\log
\frac{p(\mathbf{Z}_Y\mid\mathbf{H}_Y)}
     {p(\mathbf{Z}_Y)}
\right]
$}
\end{equation}

Since the true posterior $p(\mathbf{Z}_Y \mid \mathbf{H}_Y)$ is generally intractable, we introduce a variational encoder $q_\phi(\mathbf{Z}_Y \mid \mathbf{H}_Y)$ to approximate it. Multiplying the log-term by $\frac{q_\phi(\mathbf{Z}_Y \mid \mathbf{H}_Y)}{q_\phi(\mathbf{Z}_Y \mid \mathbf{H}_Y)}=1$ and splitting the expectation yields:

\begin{equation}
\scalebox{0.65}{$
\begin{aligned}
I(\mathbf{Z}_Y;\,\mathbf{H}_Y)
&=
\mathbb{E}_{p(\mathbf{Z}_Y,\mathbf{H}_Y)}
\!\left[
\log
\frac{
q_\phi(\mathbf{Z}_Y\mid\mathbf{H}_Y)
}{
p(\mathbf{Z}_Y)
}
\right]
\\
&\quad+
\mathbb{E}_{p(\mathbf{H}_Y)}
\!\Big[
D_{\mathrm{KL}}
\!\big(
p(\mathbf{Z}_Y\mid\mathbf{H}_Y)
\,\|\,
q_\phi(\mathbf{Z}_Y\mid\mathbf{H}_Y)
\big)
\Big].
\end{aligned}
$}
\end{equation}

By the non-negativity of the KL divergence, i.e., $D_{\mathrm{KL}}(p \parallel q_\phi) \geq 0$, dropping the second term gives the inequality:

\begin{equation}
\scalebox{0.92}{$
I(\mathbf{Z}_Y;\,\mathbf{H}_Y)
\le
\mathbb{E}_{p(\mathbf{Z}_Y,\mathbf{H}_Y)}
\!\left[
\log
\frac{
q_\phi(\mathbf{Z}_Y\mid\mathbf{H}_Y)
}{
p(\mathbf{Z}_Y)
}
\right]
$}
\end{equation}

Factorizing $p(\mathbf{Z}_Y, \mathbf{H}_Y) = p(\mathbf{H}_Y)\,p(\mathbf{Z}_Y \mid \mathbf{H}_Y)$ and using $q_\phi$ as the sampling distribution (via the reparameterization trick) for the inner expectation, we arrive at the Variational Upper Bound (VUB):

\begin{equation}
\scalebox{0.75}{$
I(\mathbf{Z}_Y;\, \mathbf{H}_Y)
\leq
\mathbb{E}_{p(\mathbf{H}_Y)}
\!\Big[
D_{\mathrm{KL}}
\!\big(
q_\phi(\mathbf{Z}_Y \mid \mathbf{H}_Y)
\parallel
p(\mathbf{Z}_Y)
\big)
\Big]
$}
\label{eq:vub}
\end{equation}

Minimizing this upper bound restricts the information capacity of $\mathbf{Z}_Y$, forcing it to discard redundant content from $\mathbf{H}_Y$.

\subsection{Ornstein--Uhlenbeck Prior and Tractable KL Divergence}
\label{app:ou_prior}
\paragraph{Choice of the OU Prior.}
Specifying the prior $p(\mathbf{Z}_Y)$ is essential for computing the VUB in Eq.~\eqref{eq:vub}, as it serves as the reference distribution that regularizes the encoder $q_\phi(\mathbf{Z}_Y \mid \mathbf{H}_Y)$.  Standard VAEs typically assume an independent standard normal prior, which ignores the temporal continuity inherent in expressive speech (e.g., prosody and emotion evolve smoothly). To capture this temporal inertia, we model the prior as a discrete-time Ornstein--Uhlenbeck (OU) process—a mean-reverting first-order autoregressive, AR(1), process. The joint prior factorizes as:

\begin{equation}
\scalebox{0.8}{$
p(\mathbf{Z}_Y)
=
\prod_{i=1}^{M}
p(\mathbf{z}_{y_i}\mid\mathbf{z}_{y_{i-1}}).
$}
\end{equation}

where
$p(\mathbf{z}_{y_i}\mid\mathbf{z}_{y_{i-1}})
=
\mathcal{N}(\alpha \mathbf{z}_{y_{i-1}}, \sigma_p^2 \mathbf{I})$.

Here, $\alpha \in [0, 1)$ controls the temporal smoothness, and $\sigma_p^2$ is the prior variance.

Assuming the variational encoder extracts the expressive intent $\mathbf{z}_{y_i}$ from a Gaussian posterior $q_\phi(\mathbf{z}_{y_i} \mid \mathbf{h}_{y_i}) = \mathcal{N}(\boldsymbol{\mu}_i, \mathrm{diag}(\boldsymbol{\sigma}_i^2))$ via the reparameterization trick:
\begin{equation}
    \mathbf{z}_{y_i} = \boldsymbol{\mu}_i + \boldsymbol{\sigma}_i \odot \boldsymbol{\epsilon}, \quad \boldsymbol{\epsilon} \sim \mathcal{N}(\mathbf{0}, \mathbf{I}),
\end{equation}
the sequence-level KL divergence admits a chain-rule decomposition:

\begin{equation}
\scalebox{0.65}{$
\begin{aligned}
&D_{\mathrm{KL}}\!\left(
q_\phi(\mathbf{Z}_Y \mid \mathbf{H}_Y)
\,\|\, p(\mathbf{Z}_Y)
\right) \\
&\quad =
\sum_{i=1}^{M}
\mathbb{E}_{q_\phi(\mathbf{z}_{y_{i-1}} \mid \mathbf{h}_{y_{i-1}})}
\Big[
D_{\mathrm{KL}}\!\left(
q_\phi(\mathbf{z}_{y_i} \mid \mathbf{h}_{y_i})
\,\|\,
p(\mathbf{z}_{y_i} \mid \mathbf{z}_{y_{i-1}})
\right)
\Big].
\end{aligned}
$}
\label{eq:kl_markov}
\end{equation}

\paragraph{Mean Substitution and the Stop-Gradient Operator.}
Assuming diagonal-Gaussian parameterizations for the posterior $q_\phi(\mathbf{z}_{y_i} \mid \mathbf{h}_{y_i})$ and the conditional prior $p(\mathbf{z}_{y_i} \mid \mathbf{z}_{y_{i-1}})$, the inner KL divergence for a $d$-dimensional latent space can be computed analytically:

\begin{equation}
\scalebox{0.70}{$
\begin{aligned}
\mathcal{L}_{\mathrm{KL}}(i)
&=
D_{\mathrm{KL}}\!\left(
q_\phi(\mathbf{z}_{y_i}\mid\mathbf{h}_{y_i})
\,\|\,
p(\mathbf{z}_{y_i}\mid\mathbf{z}_{y_{i-1}})
\right)
\\
&=
\frac{1}{2}
\sum_{k=1}^{d}
\Bigg[
\log\frac{\sigma_p^2}{\sigma_{i,k}^2}
+
\frac{
\sigma_{i,k}^2
+
(\mu_{i,k}-\alpha z_{y_{i-1},k})^2
}{
\sigma_p^2
}
-1
\Bigg]
.
\end{aligned}
$}
\end{equation}

Which is exactly equal to Eq.~\ref{eq:kl_ou}. When evaluating the outer expectation over $q_\phi(\mathbf{z}_{y_{i-1}} \mid \mathbf{h}_{y_{i-1}})$, only the squared difference term depends on $\mathbf{z}_{y_{i-1}}$:

\begin{equation}
\scalebox{1}{$
\begin{aligned}
&
\mathbb{E}_{q_\phi(
\mathbf{z}_{y_{i-1}}
\mid
\mathbf{h}_{y_{i-1}}
)}
\!\left[
(\mu_{i,k}-\alpha z_{y_{i-1},k})^2
\right]
\\
&=
(\mu_{i,k}-\alpha\mu_{i-1,k})^2
+
\alpha^2\sigma_{i-1,k}^2 .
\end{aligned}
$}
\end{equation}

For computational simplicity, we approximate this expectation via mean substitution, absorbing the scaled variance $\alpha^2\sigma_{i-1,k}^2$ into the prior variance $\sigma_p^2$. Crucially, we apply a stop-gradient operator, $\mathrm{sg}[\cdot]$, to the previous mean $\boldsymbol{\mu}_{i-1}$:
\begin{equation}
    \mathbb{E}_{q_\phi(\mathbf{z}_{y_{i-1}} \mid \mathbf{h}_{y_{i-1}})}\!\left[\alpha\,\mathbf{z}_{y_{i-1}}\right] \approx \alpha\,\mathrm{sg}[\boldsymbol{\mu}_{i-1}].
\end{equation}
Aggregating this per-step KL divergence over the sequence yields the closed-form Variational Information Bottleneck (VIB) loss: $\mathcal{L}_{\mathrm{VIB}} = \sum_{i=1}^{M} \mathcal{L}_{\mathrm{KL}}(i)$.

\subsection{Final Tractable Objective}
Combining the reconstruction lower bound (Eq.~\eqref{eq:recon}) and the intent compression upper bound (Eq.~\eqref{eq:kl_ou}), we arrive at the final tractable loss function for the VIB-driven disentanglement:
\begin{equation}
    \mathcal{L}_{\mathrm{total}} = \mathcal{L}_{\mathrm{Recon}} + \beta \, \mathcal{L}_{\mathrm{VIB}}.
\end{equation}
By minimizing $\mathcal{L}_{\mathrm{total}}$, the model is forced to encode only the temporally smooth, acoustic-essential expressive cues into $\mathbf{Z}_Y$, successfully filtering out the redundant lexical information already present in $\mathbf{E}_Y$.

\paragraph{Optimization Dynamics of the Stop-Gradient.}
Beyond simplifying the expectation, the stop-gradient $\mathrm{sg}[\boldsymbol{\mu}_{i-1}]$ is essential for preventing representation collapse. Without it, the optimizer could trivially minimize the temporal penalty $(\mu_{i,k} - \alpha\,\mu_{i-1,k})^2$ by collapsing all $\boldsymbol{\mu}_i$ to a constant vector. The $\mathrm{sg}[\cdot]$ operator forces the current state $\boldsymbol{\mu}_i$ to actively track the historical trajectory of $\boldsymbol{\mu}_{i-1}$ without retroactively altering past representations. This ensures the OU prior acts strictly as a causal low-pass filter, enforcing temporal smoothness while preserving information capacity.

\section{Experiments Details}

\subsection{Training Configuration.}
\label{app:appendix_implementation}
The learning rate of speech head is set to $1\times10^{-5}$, while the LLM backbone is updated at $1\times10^{-6}$ during Stage 2 expressive grounding. The OU-prior hyperparameters are fixed to $\alpha=0.95$ and $\sigma_p=0.5$. To stabilize VIB optimization, the KL coefficient $\beta$ is initialized to 0 during the first 10\% warm-up steps and then annealed to 0.5 following a cosine schedule. During realization-aware alignment, we sample $K=32$ candidate speech rollouts per input and use a WER threshold $\tau=0.2$ as the intelligibility constraint. Each training stage uses 8 H20 GPUs and takes approximately two days to complete.

\subsection{Dataset Construction}
\label{sec:appendix_dataset_construction}

The training corpus is built in two layers: a \textbf{text-level
emotional QA completion} layer that yields complete emotion-labeled
quadruples $(q, e_q, a, e_a)$, , where $q$ and $a$
denote the user query and assistant answer, respectively, and $e_q$ and $e_a$ denote their corresponding emotion labels,
and an \textbf{emotion-controllable TTS}
layer that synthesizes the corresponding audio.

\paragraph{Text-level emotional QA completion.}
Raw conversational sources are heterogeneous and frequently incomplete:
some entries provide only an emotion-labeled reply, others only a query,
and many provide no emotion label at all. We therefore design a set of
symmetric LLM-based completion pipelines that fill the missing side
from any observed side, producing a complete tuple
$(q, e_q, a, e_a, c)$ where $c$ is a short audio instruction consumed by
the downstream TTS. The completion is governed by three high-level
constraints---\emph{causal coherence}, \emph{closed-vocabulary emotion
labels}, and \emph{controllable style instructions}---whose detailed
definitions are given in the prompt in
Figure~\ref{fig:qa_completion_prompt}. Each call returns a structured
output, and malformed completions are discarded.

\paragraph{Emotion-controllable TTS.}
Given a complete tuple from the text layer, the missing side is
synthesized by an emotion-controllable TTS module that conditions on
both a reference voice and the audio instruction $c$, so that speaker
identity and emotional style are decoupled and independently
controllable. The audio corpus is built around two goals:
(i) \textbf{speaker coverage}, where the user query side samples
reference voices from a voice pool while the assistant answer side
uses a single fixed voice to preserve character consistency; and
(ii) \textbf{emotion alignment}, requiring the paralinguistic
realization of the synthesized speech to be consistent with the
text-side emotion label.
Under these objectives, we reuse the same TTS module to produce three
types of audio data, serving \emph{grounding},
\emph{supervised training}, and \emph{preference learning},
respectively.

\subsection{Impact of Expressive End-to-End Training on Semantic Intelligence}

\begin{table}[t]
\centering
\small
\setlength{\tabcolsep}{3pt}
\resizebox{\columnwidth}{!}{%
\begin{tabular}{lcccccccc}
\toprule
\multirow{2}{*}{\textbf{Model Name}} 
& \multicolumn{4}{c}{\textbf{Audio Understanding}} 
& \multicolumn{4}{c}{\textbf{Empathetic Response}} \\
\cmidrule(lr){2-5} \cmidrule(lr){6-9}
& \textbf{Sound} 
& \textbf{Music} 
& \textbf{Speech} 
& \textbf{Avg.} 
& \textbf{CF} 
& \textbf{RN} 
& \textbf{Col.} 
& \textbf{IR} \\
\midrule
Qwen2.5-Omni-3B 
& 30.24 & 49.25 & 39.04 & 39.51 
& 4.54 & 4.66 & \underline{4.51} & \underline{3.24} \\
Qwen2.5-Omni-7B 
& \textbf{78.10} & \textbf{65.90} & \textbf{70.60} & \textbf{71.50} 
& \textbf{4.59} & \underline{4.58} & \textbf{4.51} & \textbf{3.28} \\

\textbf{SASLM-3B (Ours)} 
& \underline{29.94} & \underline{49.25} & \underline{39.63} & \underline{39.60} 
& \underline{4.56} & \textbf{4.65} & 4.40 & 3.24 \\
\bottomrule
\end{tabular}%
}
\caption{
Comparison of different speech language models on audio understanding (MMAU) and empathetic response ability (EchoMind). \textbf{Bold} = best; \underline{underline} = second-best.
}
\label{tab:audio_empathy}
\end{table}
\label{sec:appendix_exp_of_semantic_understanding}

To verify that end-to-end training for expressive speech does not compromise semantic intelligence, we evaluate SASLM on two complementary benchmarks: EchoMind~\cite{zhou2025echomind} for empathetic dialogue quality, assessed along four dimensions—Context Fit (CF), Response Naturalness (RN), Colloquialism (Col.), and Implicit Recognition (IR)—and MMAU-v05.15.25~\cite{sakshi2024mmau} for general audio understanding, reported as single-choice accuracy across the sound, music, and speech domains.

As shown in Table~\ref{tab:audio_empathy}, SASLM achieves an MMAU average of 39.60, on par with its base model Qwen2.5-Omni-3B (39.51), indicating that our VIB-driven modulation incurs no catastrophic forgetting of audio understanding. On EchoMind, SASLM matches the 3B baseline across all four dimensions, with a marginal gain on CF (4.56 vs.\ 4.54) and parity on IR (3.24). Together, these results demonstrate that enhanced acoustic expressiveness can be attained without sacrificing—and in some aspects mildly complementing—the model's semantic and empathetic reasoning capabilities.

\subsection{Architecture Ablation: Context Grounding and Intent Modulation}

\label{sec:appendix_complete_ablation_study}
\begin{table*}[t]
\centering

\resizebox{\textwidth}{!}{
\begin{tabular}{ll cc c cccc}
\toprule
\multirow{2}{*}{\textbf{Model Type}}
  & \multicolumn{2}{c}{\textbf{Architecture Design}}
  & \multicolumn{1}{c}{\textbf{Intell.}}
  & \multicolumn{5}{c}{\textbf{Expressiveness}} \\
\cmidrule(lr){2-3}\cmidrule(lr){4-4}\cmidrule(lr){5-9}
  & $\mathcal{M}(\mathbf{e}_{y_i}, \mathbf{h}_{y_i})$
  & $\mathbf{R}_{G}$
  & \textbf{WER(\%)$\downarrow$}
  & \textbf{F0-Var$\uparrow$}
  & \textbf{Emo.$\uparrow$}
  & \textbf{Pro.$\uparrow$}
  & \textbf{Nat.$\uparrow$}
  & \textbf{Ovr.$\uparrow$} \\
\midrule

\multirow{3}{*}{Conversational TTS}
  & \multirow{3}{*}{$\pi_1$}
  & $\varnothing$    & \textbf{3.53} & 55.88 & 4.04 & 4.20 & 4.30 & 4.17 \\
  &
  & Semantic         & 9.62          & 61.16 & 4.02 & 4.18 & 4.28 & 4.16 \\
  &
  & Acoustic         & 3.61          & 60.42 & 4.06 & 4.22 & 4.30 & 4.17 \\

\midrule

\multirow{3}{*}{Naive Addition}
  & \multirow{3}{*}{$\oplus$}
  & $\varnothing$    & 7.91  & 57.00 & 4.06 & 4.22 & 4.32 & 4.19 \\
  &
  & Semantic         & 15.40 & 57.61 & 4.11 & 4.24 & 4.35 & 4.21 \\
  &
  & Acoustic         & 8.29  & 57.10 & 4.04 & 4.28 & 4.31 & 4.20 \\

\midrule

\multirow{3}{*}{Vanilla AdaLN}
  & \multirow{3}{*}{$\mathrm{AdaLN}$}
  & $\varnothing$    & 5.66  & 59.03 & 4.00 & 4.17 & 4.25 & 4.14 \\
  &
  & Semantic         & 15.13 & 61.14 & 4.02 & 4.19 & 4.27 & 4.15 \\
  &
  & Acoustic         & 6.50  & \textbf{62.24} & 4.15 & 4.28 & 4.34 & 4.22 \\

\midrule

\multirow{3}{*}{SASLM (Ours)}
  & \multirow{3}{*}{$\mathrm{VIB{+}AdaLN}$}
  & $\varnothing$    & 4.19  & 56.34 & 4.10 & 4.23 & 4.33 & 4.21 \\
  &
  & Semantic         & 11.97 & 56.43 & 4.08 & 4.24 & 4.36 & 4.20 \\
  &
  & \cellcolor{gray!15}Acoustic
  & \cellcolor{gray!15}\underline{4.74}
  & \cellcolor{gray!15}58.37
  & \cellcolor{gray!15}\textbf{4.21}
  & \cellcolor{gray!15}\textbf{4.34}
  & \cellcolor{gray!15}\textbf{4.45}
  & \cellcolor{gray!15}\textbf{4.31} \\

\bottomrule
\end{tabular}
}

\caption{
Ablation study on expressive-aware speech generation across two independent design choices in Eq.~\ref{eq:speech_generation}.
\textbf{(1) Intent Modulation Strategy} $\mathcal{M}(\mathbf{e}_{y_i}, \mathbf{h}_{y_i})$:  how semantic intent is injected for expressive generation, where
$\pi_1$: $\mathcal{M}(\mathbf{e}_{y_i}, \mathbf{h}_{y_i}) = \mathbf{e}_{y_i}$ 
discards SLM semantic intent (conversational TTS-style baseline);
$\oplus$: naive element-wise addition $\mathbf{e}_{y_i} + \mathbf{h}_{y_i}$~\cite{xu2025qwen25omnitechnicalreport};
$\mathrm{AdaLN}$~\cite{wang2025empathy}: adaptive modulation without VIB;
$\mathrm{VIB{+}AdaLN}$: our proposed disentangled modulation.
\textbf{(2) Context Grounding} $\mathbf{R}_G$: how context is constructed during synthesis, where
$\varnothing$ denotes no context grounding,
\textit{Semantic} denotes perception-pathway representations $\mathbf{R}_G = \mathbf{E}_X$, and 
\textit{Acoustic} denotes generation-encoder acoustic features $\mathbf{R}_G = \phi_G(X^S)$.
}
\label{tab:ablation_fusion_full}

\end{table*}

Table~\ref{tab:ablation_fusion_full} reports the complete ablation results across all combinations of fusion strategy and context grounding after the first two training stages, extending the summary in Section~\ref{sec:ablation_study_architecture}.

\paragraph{Intent Modulation Strategy.}
Table~\ref{tab:ablation_fusion_full} isolates how semantic intent is injected into speech generation under acoustic grounding. The conversational TTS baseline ($\pi_1$) attains strong intelligibility (WER 3.61\%) but limited expressiveness (Ovr.~4.17), confirming the absence of a pathway from \textit{what the model thinks} to \textit{how it speaks}. \textbf{Naive injection harms intelligibility with little expressive gain.} Direct summation ($\oplus$) entangles fast-varying lexical content with token-level acoustics, inflating WER to 8.29\% while barely improving expressiveness (Ovr.~4.20). Vanilla AdaLN treats $\mathbf{h}_{y_i}$ as a modulation signal rather than an additive perturbation, reducing WER to 6.50\%, yet lexical information still leaks without regularization. \textbf{VIB regularization resolves intent transmission failure.} VIB+AdaLN compresses $\mathbf{h}_{y_i}$ into a temporally smooth intent variable $\mathbf{z}_{y_i}$ via an OU prior, suppressing lexical fluctuations while preserving utterance-level intent. This yields the best expressiveness (Ovr.~4.31, Emo.~4.21, Pro.~4.34, Nat.~4.45) with near-baseline intelligibility (WER 4.74\%), confirming that disentangling intent from content is essential for effective expressive modulation.

\paragraph{Context Grounding.}
We next vary the grounding signal $\mathbf{R}_G$ across all modulation strategies. Removing context ($\varnothing$) improves intelligibility but weakens expressiveness: under VIB+AdaLN, $\varnothing$ yields WER 4.19\%/Ovr.4.21 versus 4.74\%/4.31 with acoustic grounding, indicating that acoustic context contributes to both pronunciation stability and prosodic variation. \textbf{Semantic grounding is incompatible with reliable synthesis.} Replacing acoustic features with perception-pathway representations ($\mathbf{R}_G = \mathbf{E}_X$) yields the worst WER across all strategies (9.62\%--15.40\%), as such representations lack the fine-grained acoustic continuity needed for phoneme-level articulation. \textbf{Acoustic grounding is indispensable for jointly preserving intelligibility and expressiveness.} $\phi_G(X^S)$ supplies the speech head with continuous acoustic representations that retain phonetic detail while remaining decoupled from the LLM semantic pathway, better capturing the one-to-many nature of spoken interaction\cite{WynnB22}.

\section{Evaluation Details}
\subsection{Prompt Design for Conversation-level Subjective Evaluation}
\label{app:appendix_prompt}

We employ the Gemini-2.5-Flash API for conversation-level subjective evaluation along three dimensions: \textit{emotion and intensity}, \textit{prosody}, and \textit{naturalness}. All three prompts follow a unified two-stage protocol, with the emotion-and-intensity prompt shown in Fig.~\ref{fig:emotion_intensity_prompt} as a representative example. The model first infers the expected delivery (target emotion and intensity) from the dialogue context and response text, then listens to the synthesized audio to judge whether the actual delivery matches this expectation. By decoupling \textit{expected expression} from \textit{acoustic execution}, this design promotes context-grounded judgments that go beyond surface acoustic quality to assess whether the speech faithfully conveys the dialogue's communicative intent.

\subsection{Human Evaluation}
\label{app:human_evaluation}

We recruited three annotators based on two criteria: (1) an IELTS score
of 6.5 or above, ensuring sufficient English listening ability to judge
fine-grained expressive cues; and (2) a computer-science background
with research experience or basic research training, helping them
understand the evaluation criteria and produce consistent judgments. 
During annotation, each audio clip is presented together with its
subtitle so that annotators can focus on acoustic expressiveness
rather than being affected by speech-recognition difficulty.All reported statistics are aggregated across the three annotators.

\subsubsection{Speech Performance Assessment}
\label{app:speech_assessment}

\begin{figure}[t]
\centering
\includegraphics[width=\linewidth]{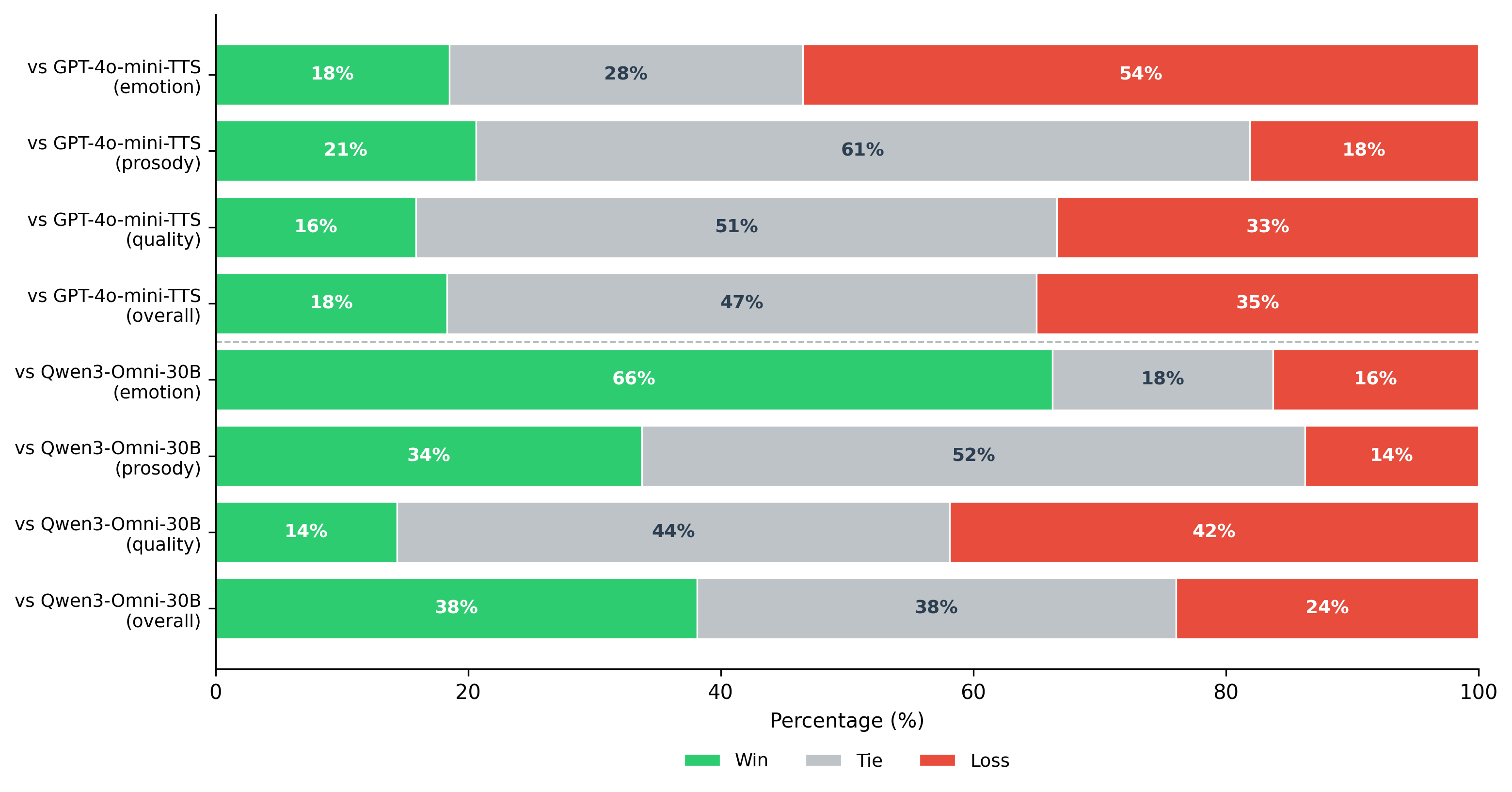}
\caption{Pairwise human Win/Tie/Loss between our model and each
baseline (GPT-4o-mini-TTS, Qwen3-Omni-30B) on the three perceptual
dimensions, together with an aggregated \emph{Overall} row per
baseline. \emph{Win} (green) indicates that our model is preferred,
\emph{Loss} (red) indicates that the baseline is preferred, and
\emph{Tie} (gray) indicates perceptually comparable.}
\label{fig:mos_human_eval}
\end{figure}

To assess perceptual speech performance, we conduct a pairwise human
comparison study. For each utterance, annotators independently judge
our model against each baseline along three perceptual dimensions --
\emph{Emotion}, \emph{Prosody}, and \emph{Quality} -- yielding a
\emph{Win}, \emph{Loss}, or \emph{Tie} verdict per (annotator,
utterance, dimension) triplet; an aggregated \emph{Overall} verdict
is summarized as the average of the three per-dimension Win/Tie/Loss
rates. Figure~\ref{fig:mos_human_eval} reports the aggregated
Win/Tie/Loss distributions, from which we observe that our model is
clearly preferred over the strong open-source baseline
Qwen3-Omni-30B (overall Net Win Rate $+14.2\%$) but trails the
proprietary GPT-4o-mini-TTS (overall Net Win Rate $-16.7\%$); the
remaining \emph{Quality} deficit against both baselines is mainly
caused by occasional acoustic artifacts (e.g.\ mild breaking,
breathiness, unstable energy), which we attribute to the limited
scale of our training data and which also explain the slightly
higher WER of our model in the main results.

\subsubsection{Reward-Model Validation}
\label{app:reward_validation}

\begin{figure}[t]
\centering
\includegraphics[width=\linewidth]{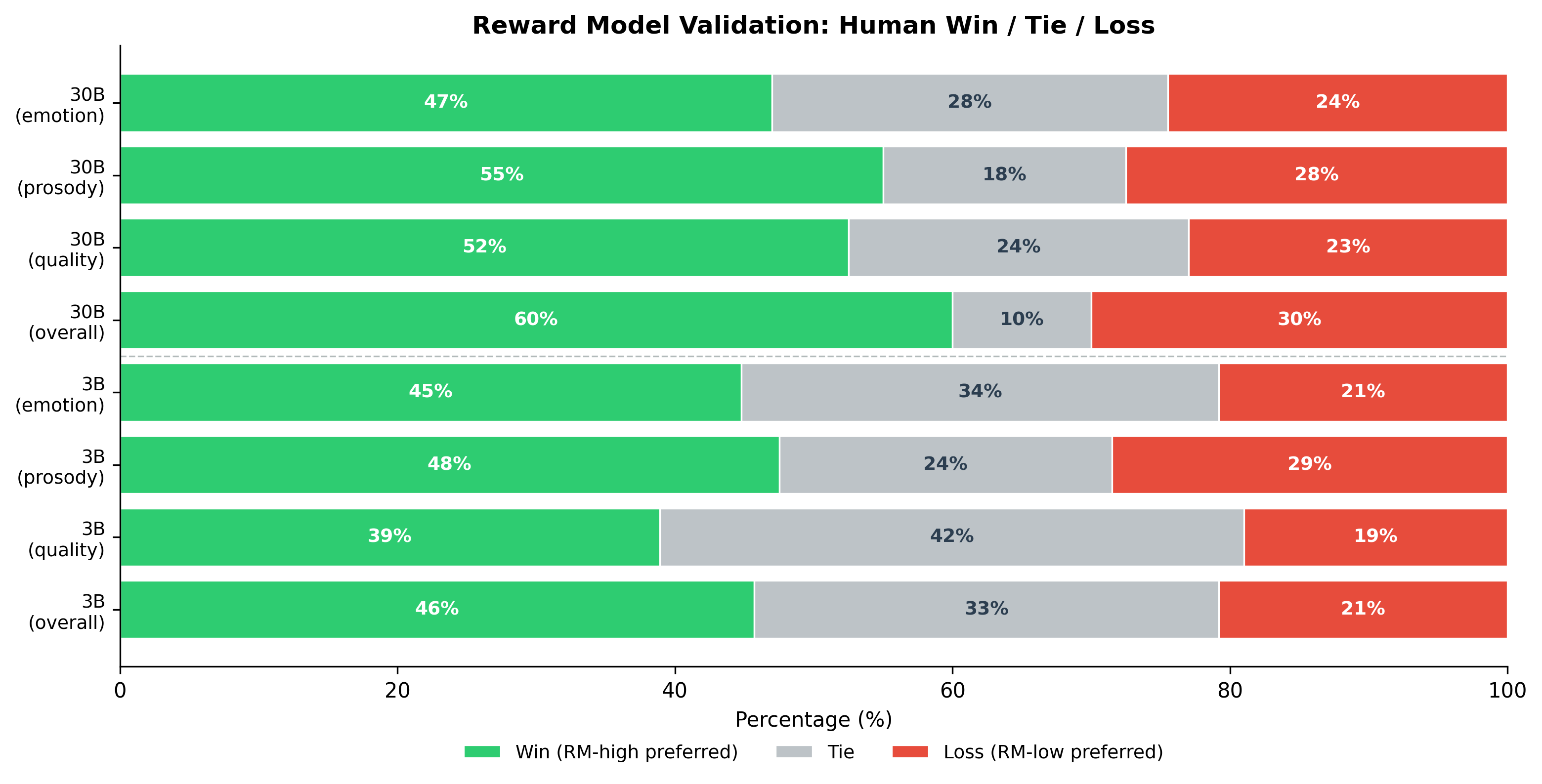}
\caption{Human Win/Tie/Loss validation of reward models. Each row
corresponds to a (model, dimension) configuration. \emph{Win}
(green) indicates that the RM-high sample is preferred by human
raters, \emph{Loss} (red) indicates that the RM-low sample is
preferred, and \emph{Tie} (gray) indicates that the two samples are
judged perceptually comparable.}
\label{fig:reward_model_validation}
\end{figure}

To verify whether reward-model preferences are aligned with human
judgments, we conduct a blind pairwise comparison study. For each
pair, the reward model assigns a higher score to one sample
(RM-high) and a lower score to the other (RM-low). The two samples
are then presented to annotators in a randomized order, while their
RM-induced ranking is hidden, and annotators compare them on
\emph{Emotion}, \emph{Prosody}, and \emph{Quality} as a Win, Loss,
or Tie. As shown in Figure~\ref{fig:reward_model_validation}, the
reward-model-selected samples are generally preferred by human
annotators, indicating a positive alignment between reward-model
rankings and human perceptual judgments.

\section{Prompt Template}
\begin{figure*}[t]
\begin{tcolorbox}[
    colback=gray!3, 
    colframe=black!75, 
    title=\textbf{Prompt for Emotion Delivery Evaluation},
    fonttitle=\bfseries,
    boxrule=0.6pt,
    arc=2pt,
    left=6pt, right=6pt, top=4pt, bottom=4pt
]

\textbf{Role} \\
You are a professional \emph{Speech Quality Reviewer}. Your task is to assess whether a synthesized speech faithfully realizes the specified emotional directive, based strictly on its acoustic delivery.

\medskip
\textbf{Inputs}
\begin{itemize}[leftmargin=1.2em, itemsep=1pt, topsep=2pt, parsep=0pt]
    \item \textbf{Emotion Directive}: The intended emotion and arousal level.
    \item \textbf{Target Transcript}: The textual content to be spoken.
    \item \textbf{Response Speech}: The synthesized audio to be evaluated.
\end{itemize}

\medskip
\textbf{Core Principles}
\begin{itemize}[leftmargin=1.2em, itemsep=1pt, topsep=2pt, parsep=0pt]
    \item \textbf{Acoustic-Grounded}: Judgments must rely solely on \emph{acoustic features} (pitch, intonation, rate, energy), \textbf{not} on textual semantics.
    \item \textbf{Anti-Exaggeration}: If the delivery sounds flat, describe it honestly. \textbf{Never} project emotions onto the speech to satisfy the directive.
    \item \textbf{Objective Description}: Specify both the \emph{perceived emotion type} and the \emph{arousal level} (low / medium / high).
\end{itemize}

\medskip
\textbf{Scoring Rubric} (\texttt{status} $\in \{0, 0.5, 1\}$)
\begin{itemize}[leftmargin=1.2em, itemsep=1pt, topsep=2pt, parsep=0pt]
    \item \textbf{1 — Match}: Delivery fully aligns with the directive.
    \item \textbf{0.5 — Partial}: Target cues present but intensity insufficient (e.g., \emph{rage} directive delivered as \emph{mild annoyance}).
    \item \textbf{0 — Mismatch}: Delivery clearly contradicts the directive (e.g., \emph{sad} read \emph{cheerfully}).
\end{itemize}

\medskip
\textbf{Output Format}
\begin{verbatim}
{
  "check_emotion_directive": {
    "instruction": "<copy of emotion_directive, verbatim>",
    "observation": "<objective acoustic-based description>",
    "status": <0 | 0.5 | 1>
  }
}
\end{verbatim}

\end{tcolorbox}
\caption{Prompt for evaluating emotion delivery in synthesized speech.}
\label{fig:emotion_rl_prompt}
\end{figure*}
\begin{figure*}[t]
\begin{tcolorbox}[colback=white, colframe=black,
  title=Prompt for Text-level Emotional QA Completion]
\textbf{Role.} You are a dialogue designer crafting emotionally
engaging, natural conversations.

\textbf{Task.} Given one side of a dialogue (e.g., Character B's
emotion-labeled reply), generate the missing counterpart
(Character A's preceding line) with its emotion label and an audio
instruction specifying how it should be spoken.

\medskip
\textbf{(1) Causal coherence.}
The generated line must serve as a natural emotional cause of the
observed side, such that its removal would render the reply
incomplete. The model must justify this causality in a
\texttt{causality\_check} field, avoiding generic reactions
(e.g., ``What happened?'') and lexical repetition of the observed side.

\medskip
\textbf{(2) Closed-vocabulary emotion labels.}
The emotion must be \emph{a single word} drawn \emph{verbatim} from a
predefined closed vocabulary of 20 labels, covering 14 primary
emotions (e.g., Happy, Sad, Angry, \ldots) and 6 extended ones
(e.g., Gentle, Playful, \ldots). Compound labels, modifiers
(e.g., ``Very''), and out-of-vocabulary terms are forbidden.

\medskip
\textbf{(3) Controllable style instruction.}
The audio instruction is a 3--8 word descriptor ending with
\texttt{<|endofprompt|>}, conveying exactly one emotion at one of two
intensity levels (\emph{Normal} or \emph{High}). Softening modifiers
(e.g., \emph{slightly}, \emph{mildly}), multi-emotion combinations,
and indirect framings (e.g., ``Barely holding back anger'') are
forbidden. Example: ``Strong sorrowful tone\texttt{<|endofprompt|>}''.
\end{tcolorbox}
\caption{Prompt for text-level emotional QA completion. The figure
illustrates the ``answer-to-question'' direction; other directions
follow the same template by swapping the observed and generated sides.}
\label{fig:qa_completion_prompt}
\end{figure*}
\begin{figure*}[t]
\begin{tcolorbox}[
    colback=gray!3, 
    colframe=black!75, 
    title=\textbf{Prompt for Emotion and Intensity Evaluation},
    fonttitle=\bfseries,
    boxrule=0.6pt,
    arc=2pt,
    left=6pt, right=6pt, top=4pt, bottom=4pt
]

\textbf{Objective} \\
Given a conversational context, a response text, and its synthesized speech, evaluate whether the speech delivery is emotionally appropriate and contextually calibrated.

\medskip
\textbf{Inputs}
\begin{itemize}[leftmargin=1.2em, itemsep=1pt, topsep=2pt, parsep=0pt]
    \item \textbf{Speech Input}: User's utterance providing conversational grounding.
    \item \textbf{Response Text}: Target textual content to be synthesized.
    \item \textbf{Response Speech}: Generated speech of the response.
\end{itemize}

\medskip
\textbf{Procedure}
\begin{enumerate}[leftmargin=1.5em, itemsep=1pt, topsep=2pt, parsep=0pt]
    \item \textbf{Expectation}: Infer the target emotion and intensity (e.g., empathetic, encouraging, calm, enthusiastic, neutral) from the input and response text \emph{before} listening.
    \item \textbf{Execution}: Listen to the response speech and compare actual delivery with expectation, checking for tonal appropriateness and intensity calibration.
\end{enumerate}

\medskip
\textbf{Scoring Rubric}
\begin{itemize}[leftmargin=1.2em, itemsep=1pt, topsep=2pt, parsep=0pt]
    \item \textbf{5 — Perfect}: Precise, evocative, perfectly calibrated.
    \item \textbf{4 — High}: Well aligned; minor issues in subtle transitions.
    \item \textbf{3 — Acceptable}: Basic tendency conveyed but generic or mildly mis-scaled.
    \item \textbf{2 — Poor}: Clearly inconsistent or substantially miscalibrated.
    \item \textbf{1 — Unacceptable}: Flat, inappropriate, or uncomfortable delivery.
\end{itemize}

\medskip
\textbf{Output Format}
\begin{verbatim}
{
  "Expected_Emotion_and_Intensity": "...",
  "Execution_Analysis": "...",
  "Score": <1-5>
}
\end{verbatim}

\end{tcolorbox}
\caption{Prompt for evaluating emotion and intensity in synthesized speech.}
\label{fig:emotion_intensity_prompt}
\end{figure*}
\clearpage
\onecolumn
\begin{figure}[t]
\centering
\begin{tcolorbox}[
    colback=gray!5,
    colframe=black,
    fonttitle=\bfseries,
    title=Guidelines for Human MOS Annotation,
    width=\linewidth,
    boxrule=0.5pt,
    arc=1pt,
]
\textbf{Task.}
You will listen to a short speech utterance generated by a speech
model in a multi-turn dialogue. The dialogue subtitle is provided.
Please rate the audio on a 1--5 scale along the three perceptual
dimensions below, and additionally provide a single \emph{Overall}
score that summarizes your holistic preference, where $1$ = very
poor and $5$ = excellent.

\medskip
\textbf{Dimensions.}
\begin{itemize}\itemsep0pt
    \item \textbf{Emotion.} Is the emotional delivery appropriate
    for the content? Does it match what a speaker would naturally
    feel in this context?
    \item \textbf{Prosody.} Is the pitch, rhythm, stress and pause
    pattern rich and natural? Is the speech free of monotone or
    robotic delivery?
    \item \textbf{Quality.} Is the audio signal clean? Listen for
    glitches, distortion, breaking, breathiness, and unstable
    energy.
    \item \textbf{Overall.} A holistic judgment of how good this
    audio is as a response in this conversation, considering
    expressiveness, naturalness and clarity together.
\end{itemize}

\medskip
\textbf{Procedure (MOS study).}
Listen to each audio at least once with the subtitle visible, and
provide one integer rating per dimension.

\medskip
\textbf{Procedure (reward-model validation).}

The ordering of samples is randomized and
the reward-model ranking is hidden from
annotators.
Two audio samples are presented as a pair.
Considering the same perceptual criteria
(Emotion, Prosody, and Quality), select

\begin{itemize}
\item \textbf{Win}: RM-high is preferred;
\item \textbf{Loss}: RM-low is preferred;
\item \textbf{Tie}: no clear perceptual preference.
\end{itemize}
\end{tcolorbox}
\caption{Annotation guideline shown to human evaluators in both the
MOS study and the reward-model validation study.}
\label{fig:human_eval_prompt}
\end{figure}



\end{document}